\documentclass{article}

     \usepackage[final, nonatbib]{neurips_2019}

\usepackage[utf8]{inputenc} 
\usepackage[T1]{fontenc}    
\usepackage{hyperref}       
\usepackage{url}            
\usepackage{booktabs}       
\usepackage{amsfonts}       
\usepackage{nicefrac}       
\usepackage{microtype}      
\usepackage[pdftex]{graphicx}
\usepackage{svg}            
\usepackage{soul, xcolor}   
\usepackage{siunitx}        
\usepackage{float}          
\usepackage{listings}       
\usepackage{tcolorbox}      
\usepackage{tablefootnote}  

\newcommand{\ctext}[3][RGB]{%
  \begingroup
  \definecolor{hlcolor}{#1}{#2}\sethlcolor{hlcolor}%
  \hl{#3}%
  \endgroup
}

\title{GPT-Sentinel: Distinguishing Human and ChatGPT Generated Content}

\author{
Yutian Chen$^\dagger$\\
School of Computer Science\\
Carnegie Mellon University\\
Pittsburgh, PA 15213 \\
\texttt{yutianch@andrew.cmu.edu}\\
\And
Hao Kang$^\dagger$\\
School of Computer Science\\
Carnegie Mellon University\\
Pittsburgh, PA 15213\\
\texttt{haok@andrew.cmu.edu}
\And
Vivian Zhai$^\dagger$\\
College of Engineering\\
Carnegie Mellon University\\
Pittsburgh, PA 15213\\
\texttt{yiyanz@andrew.cmu.edu}
\AND
Liangze Li\\
Language Technologies Institute\\
Carnegie Mellon University\\
Pittsburgh, PA 15213\\
\texttt{liangzel@andrew.cmu.edu}
\And
Rita Singh\\
Language Technologies Institute\\
Carnegie Mellon University\\
Pittsburgh, PA 15213\\
\texttt{rsingh@cs.cmu.edu}
\And
Bhiksha Raj\\
Language Technologies Institute\\
Carnegie Mellon University\\
Pittsburgh, PA 15213\\
\texttt{bhiksha@cs.cmu.edu}
}

\begin{document}

\maketitle
\def\thefootnote{$\dagger$}
\footnotetext{Three authors contribute equally to this work.}
\def\thefootnote{\arabic{footnote}}

\begin{abstract}
  This paper presents a novel approach for detecting ChatGPT-generated vs. human-written text using language models. To this end, we first collected and released a pre-processed dataset named \texttt{OpenGPTText}, which consists of rephrased content generated using ChatGPT. We then designed, implemented, and trained two different models for text classification, using Robustly Optimized BERT Pretraining Approach (RoBERTa) and Text-to-Text Transfer Transformer (T5), respectively. Our models achieved remarkable results, with an accuracy of over 97\% on the test dataset, as evaluated through various metrics. Furthermore, we conducted an interpretability study to showcase our model's ability to extract and differentiate key features between human-written and ChatGPT-generated text. Our findings provide important insights into the effective use of language models to detect generated text.

\end{abstract}

\section{Introduction}
The development of an algorithm that can accurately distinguish between machine-generated text and human-generated text has become crucial in contexts where verifying the authenticity of information is essential, such as in legal proceedings and news reporting. Although traditional statistical techniques such as logistic regression and support vector machines (SVM) have been used for this purpose in the past \cite{DBLP:journals/corr/abs-2011-01314}, the emergence of Large Language Models (LLMs) like InstructGPT \cite{InstructGPT} and the availability of its free deployment, ChatGPT, has presented significant challenges to existing detection methods. As a result, the need to develop novel algorithms that can accurately distinguish between machine and human-generated text has become more pressing than ever before.

To address this issue, we focused on fine-tuning approaches to distinguish human-written and ChatGPT-generated text. We first collected the data from ChatGPT and established the \texttt{OpenGPTText} data set. Section 3 of the paper provides a detailed description of the data collection process, including the criteria to select the samples and the methods to filter out irrelevant and undesired noise in the collected text. We then trained the frozen RoBERTa with MLP and fine-tuned the T5 model on this data set for classification. The resulting model is what we referred to as \textit{GPT-Sentinel}. More details about the model can be found in Section 4 of the paper.

The rest of this paper is structured as follows: We first discuss related work in Section 2; illustrate \texttt{OpenGPTText} data set in Section 3; present our model and the training details in Section 4; evaluate the performance using various metrics in Section 5; interpret the basis for the model's prediction in Section 6; point out future work in Section 7; and conclude in Section 8.

\section{Related Work}
The work by Jawahar et al. identified five key characteristics that a state-of-the-art detector for content generated by LLMs should possess: \textit{accuracy}, \textit{data efficiency}, \textit{generalizability}, and \textit{interpretability} \cite{Jawahar2020}; where \textit{accuracy} means the model should be able to distinguish between LLM-generated and human-written text while achieving an appropriate trade-off between precision and recall rates; \textit{data efficiency} means that the detector should be able to operate with as few examples as possible from the language model; \textit{generalizability} means that the detector should be able to work consistently, regardless of any change in the model architecture, prompt length, or training dataset; \textit{interpretability} means the detector should provide clear explanations for the reasoning behind its decisions. These five principles is used as our guidance when designing the GPT-Sentinel.

Approaches to machine-generated text detection can be divided into three categories: traditional statistical approach (by analyzing statistical abnormality in text sample), unsupervised-learning approach (by zero-shot classification of LLM), and supervised-learning approach (by fine-tuning a language model with or without a classification module attached).  

\subsection{Statistical Methods}
The first approach to the problem is via the use of statistics. For instance, the work by Solaiman et al. \cite{Solaiman2019} demonstrated that using a logistic regression model to differentiate between text generated by GPT-2 models vs. text written by humans could achieve an accuracy ranging from 88\% (on 124 million parameter variants of GPT-2 model) to 74\% (on 1.5 billion parameter variants of GPT-2 model). Moreover, the work by Ippolito et al. \cite{Ippolito2019AutomaticDO} showed that the top-$k$ sampling method used in popular LLMs could over-sample high-likelihood words, and thus the generated text exhibited statistical anomalies, which could be further used for detection. Moreover, the statistical methods called the Giant Language Model Test Room (GLTR) designed by Gehrmann et al. \cite{DBLP:journals/corr/abs-1906-04043} consists of three tests: Tests 1 and 2 checked if a generated word is sampled from the top of the distribution, and Test 3 verified if the system is overly confident in its next prediction due to familiarity with the previously generated context. A human-subject study found that GLTR improved the human detection rate of fake text from 54\% to 72\% without prior training.

\subsection{Zero-Shot Classification}
The second detection approach is by zero-shot classification (i.e., using a pre-trained LLM to detect its own generation or that of a similar model). In the work by Solaiman et al. \cite{Solaiman2019}, a baseline method that used an LLM to evaluate the log-probability and the corresponding threshold for making classification decisions was proposed. However, this zero-shot approach performs poorly compared to the statistical methods. 

\subsection{Fine-Tuning Language Model}
The last approach is to fine-tune an existing language model. For example, Zeller et al. \cite{Defending_fake_news} fine-tuned a linear layer to identify if the input was generated by the GROVER model or by a human, using the hidden states in the encoder of GROVER. Also, Solaiman et al. \cite{DBLP:journals/corr/abs-2011-01314, Solaiman2019} fine-tuned a pre-trained RoBERTa model on a labeled dataset to create a content detector that achieved state-of-the-art performance of 90\% accuracy in detecting text generated by GPT-2. However, the supervised learning method requires a large amount of labeled data, in contrast to previously discussed methods. The fine-tuned model on RoBERTa by Solaiman et al. required 200k labeled training data.

\section{Data Set Collection}
ChatGPT is a language model based on the GPT-3.5 architecture. It succeeded InstructGPT, which was previously published by Ouyang et al. \cite{InstructGPT}. As it was introduced by OpenAI on November 30, 2022, there is currently no publicly available data set that systematically collects the outputs generated by ChatGPT as far as we know. Consequently, we undertook the task of creating our own data set for ChatGPT outputs. Building upon the work of Gokaslan et al. \cite{Gokaslan2019OpenWeb} and their \texttt{OpenWebText} corpus. We named the data set \texttt{OpenGPTText}.

\subsection{OpenGPTText Overview}

The \texttt{OpenGPTText} data set consists of paraphrased textual samples that were generated by the \texttt{gpt-3.5-turbo} language model using the \texttt{OpenWebText} corpus as its source. The data set contains 29,395 textual samples, each corresponding to a piece human-written text from the \texttt{OpenWebText} corpus that shares a same unique identifier (UID). 

Up to April 24, 2023, the \texttt{OpenGPTText} only contains approximately 1\% of paraphrased samples of the \texttt{OpenWebText} data set in some specific subsets. The number of samples in each subset is listed in table \ref{tab:OpenGPTText dataset}.

\begin{table}
    \caption{Detailed statistics for \texttt{OpenGPTText} data set as of Apr 24, 2023. The subsets not listed in the table were not paraphrased in \texttt{OpenGPTText}. The category ``Failed to Rephrase'' corresponds to one of the following situations: 1. the content length exceeds the API limit, 2. the content is blocked by OpenAI content filter.}
    \centering
    \begin{tabular}{lcccc}
    \toprule
    Subset & \texttt{OpenGPTText} & \texttt{OpenWebText} & Failed to Rephrase & Percentage \\
    \midrule
    \texttt{urlsf\_00} & $3,888$ & $391,590$ & $27 $ & $0.99\%$ \\
    \texttt{urlsf\_01} & $3,923$ & $392,347$ & $0  $ & $1.00\%$ \\
    \texttt{urlsf\_02} & $3,260$ & $391,274$ & $652$ & $0.83\%$ \\
    \texttt{urlsf\_03} & $3,891$ & $390,161$ & $10 $ & $1.00\%$ \\
    \texttt{urlsf\_04} & $3,684$ & $390,250$ & $218$ & $0.94\%$ \\
    \texttt{urlsf\_05} & $3,602$ & $389,874$ & $296$ & $0.92\%$ \\
    \texttt{urlsf\_06} & $3,494$ & $390,339$ & $409$ & $0.90\%$ \\
    \texttt{urlsf\_09} & $3,653$ & $389,634$ & $243$ & $0.94\%$ \\
    \midrule
    Total              & $29,395$ & $3,125,469$ & $1,885$ & $0.94\%$ \\
    \bottomrule
    \end{tabular}
    \label{tab:OpenGPTText dataset}
\end{table}

\subsection{Data Source}
The \texttt{OpenWebText} data set \cite{Gokaslan2019OpenWeb} is a publicly available resource that comprises web content sourced from URLs shared on Reddit with a minimum of three votes. This data set is a reconstitution of the original \texttt{WebText} corpus, which was initially described by Radford et al. \cite{radford2019language}. Since the data set was compiled in 2019, it is improbable that the textual content it contains was algorithmically generated.

\subsection{Data Collection Method}

The rephrasing procedure used OpenAI's API on \texttt{gpt-3.5-turbo} model, with the prompted instruction: ``\texttt{Rephrase the following paragraph by paragraph}''. However, it should be noted that the samples with length larger than 2,000 words were filtered out as the \texttt{gpt-3.5-turbo} can only take in at most 3,000 tokens. Some text samples blocked by OpenAI content filter was also excluded from \texttt{OpenGPTText}. The number of texts that were not successfully paraphrased due to either of the two reasons is reported in the ``Failed to Rephrase'' column in table \ref{tab:OpenGPTText dataset}.
\subsection{Data Set Cleaning}

Upon inspecting the \texttt{OpenGPTText} data set, we observed certain stylistic disparities between ChatGPT's output and the corpus in \texttt{OpenWebText}. Specifically, our analysis revealed that ChatGPT's output tend to include the Unicode character ``right double quotation mark'' (U+201D) in place of the ASCII character ``quotation mark'' (U+0022) used in the \texttt{OpenWebText} corpus. Furthermore, ChatGPT also tends to incorporate two consecutive new-line characters between paragraphs, whereas the \texttt{OpenWebText} corpus utilizes two to six new-line characters consecutively.

\begin{figure}
    \centering
    \includegraphics[width=.49\textwidth]{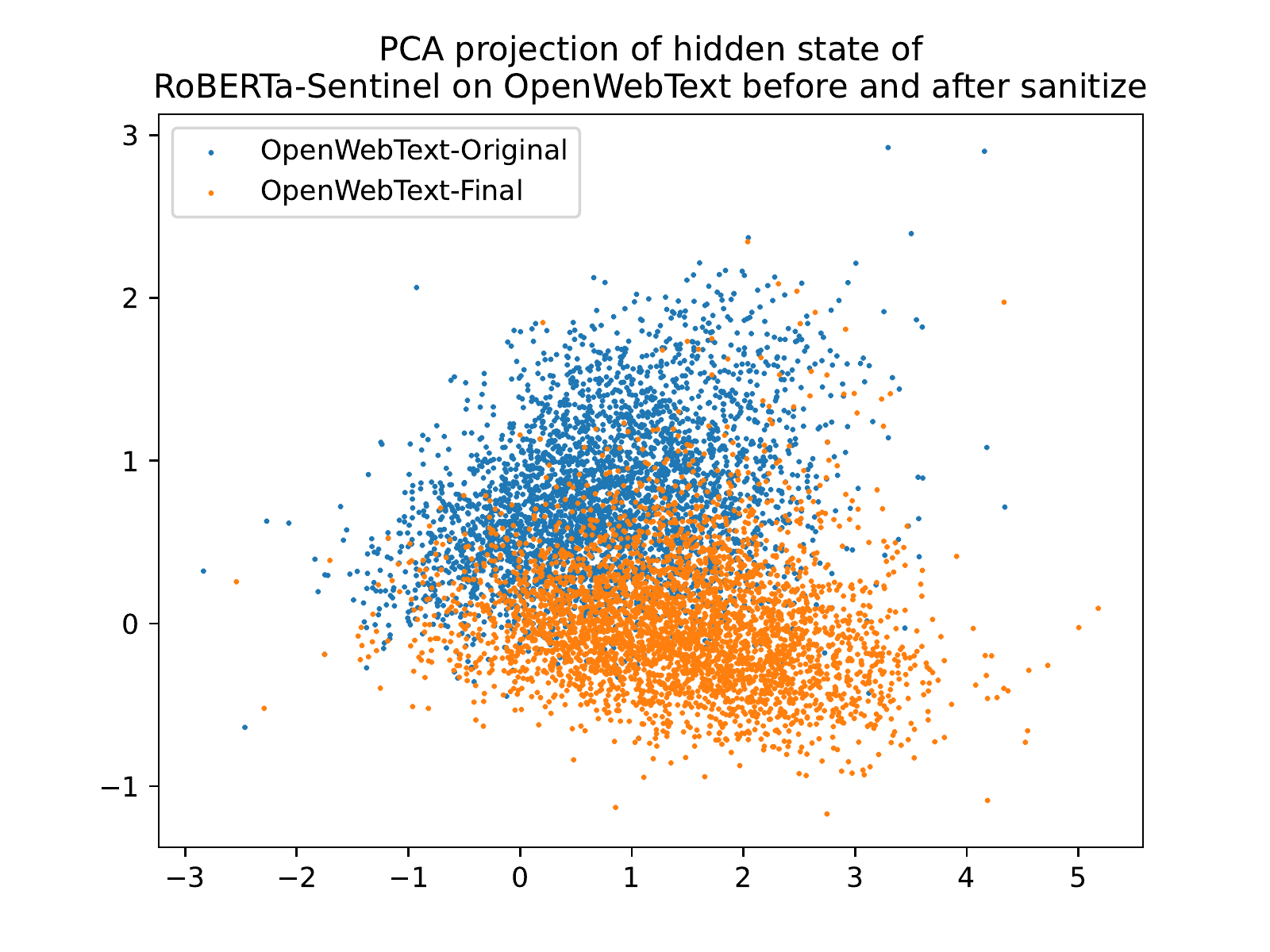}
    \includegraphics[width=.49\textwidth]{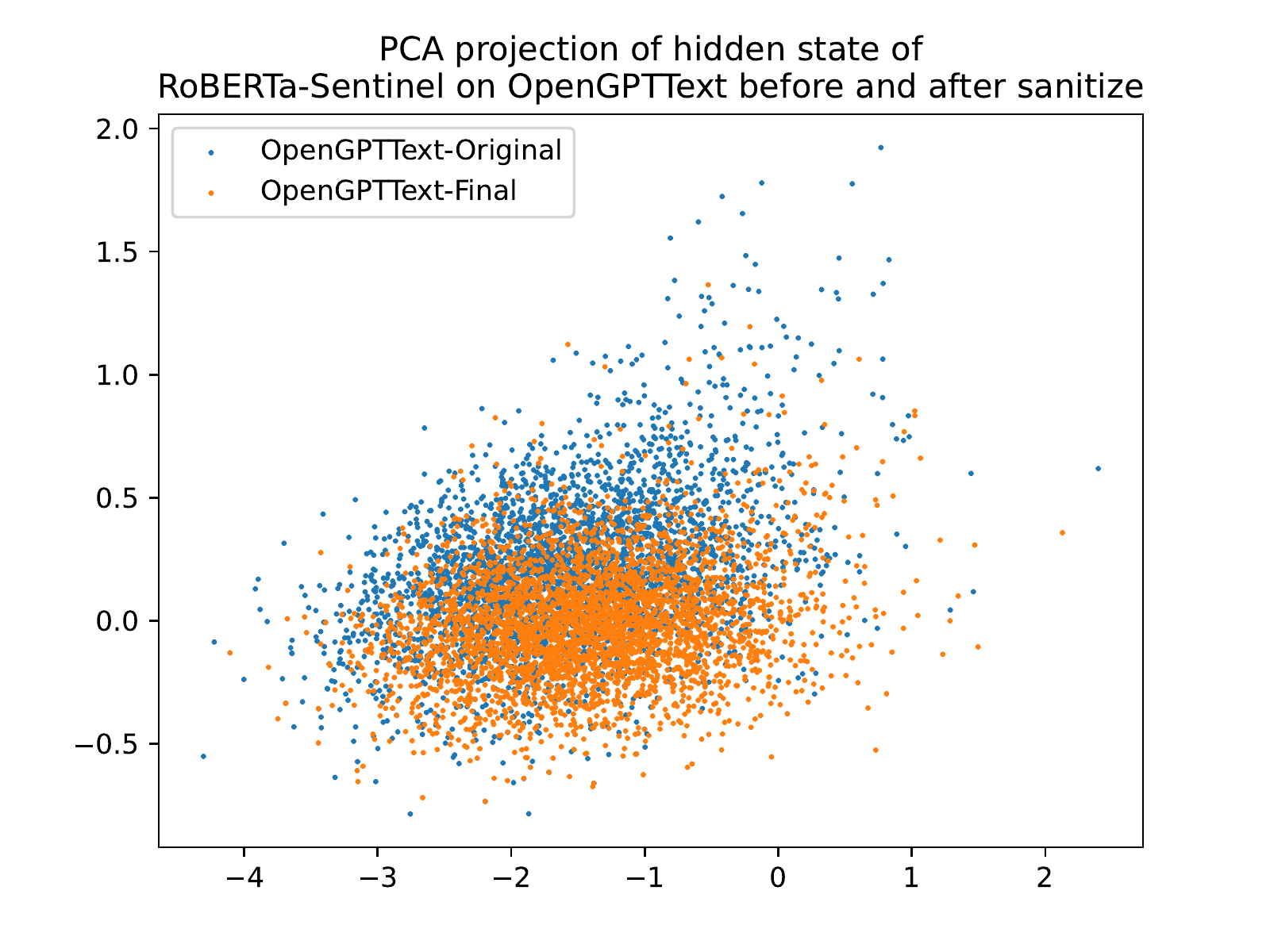}
    \caption{PCA of hidden state distribution of RoBERTa-Sentinel model on \texttt{OpenWebText} (Left) and \texttt{OpenGPTText} (Right) before and after cleaning. Note that the cleaning process affected the distribution of \texttt{OpenWebText} significantly.}
    \label{fig:PCA of dataset before and after sanitization}
\end{figure}

In an effort to enhance the resilience of our classifier and eliminate the potential influence of these susceptible features, we undertook measures to sanitize both the \texttt{OpenWebText} and \texttt{OpenGPTText} data sets. To achieve this, we implemented a cleaning procedure that involved removing excessive new-line characters and mapping Unicode characters onto the ASCII character set. These steps were taken to mitigate any possible confounding effects of these variables on the performance of our classifier. Pincipal Component Analysis (PCA) of hidden state distribution in figure \ref{fig:PCA of dataset before and after sanitization} shows that the cleaning process has significantly changed the distribution of dataset.

The resulting, clean data set are called \texttt{OpenWebText-Final} and \texttt{OpenGPTText-Final} in the discussion below.

\subsection{Data Set Release}

Our plan entails the release of both \texttt{OpenGPTText} and \texttt{OpenGPTText-Final} on Kaggle in May 2023.

\section{Method}
The following models were trained using the \texttt{OpenWebText-Final} and \texttt{OpenGPTText-Final} data set, partitioning 80\% of the data set for training, 10\% for validation, and the remaining 10\% for testing. Given that the texts in the data set have varying lengths, we truncated input text to a maximum of 512 tokens to improve training efficiency, while also padding any text with less than 512 tokens with additional \texttt{<PAD>} tokens. To address memory constraints while using a relatively large batch size during fine-tuning, we performed gradient accumulation, whereby the optimizer was updated after a certain number of forward passes.

\subsection{RoBERTa-Sentinel Model}

\begin{figure}
    \centering
    \includegraphics[width=.8\textwidth]{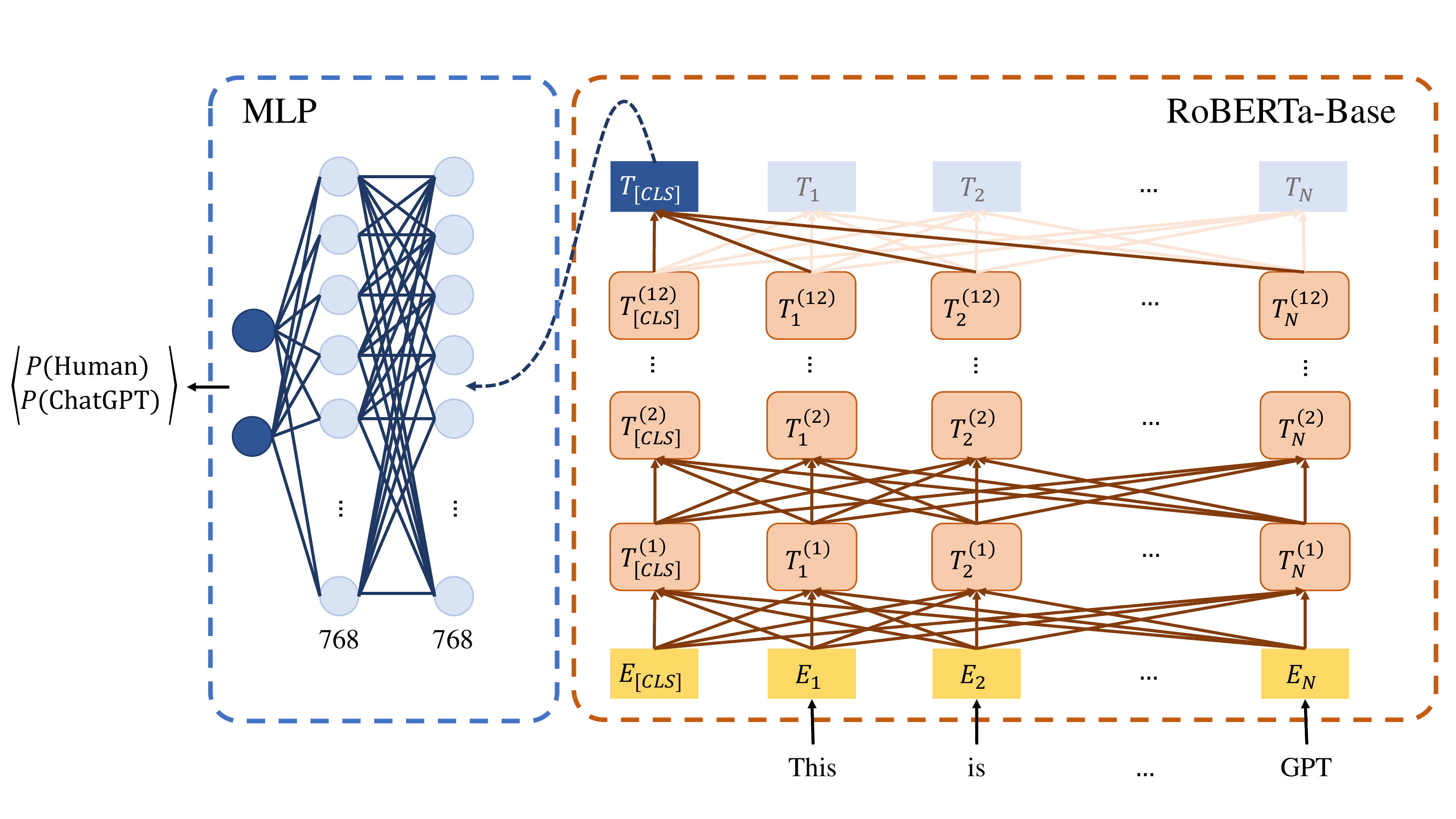}
    \caption{The depicted figure illustrates the RoBERTa-Sentinel architecture, wherein the dashed line connecting RoBERTa-Base and MLP module indicates the non-propagation of gradient back to the former.}
    \label{fig:RoBERTa Architecture}
\end{figure}

The first method we proposed is to leverage the pretrained RoBERTa model \cite{liu2019roberta} to extract relevant features from the input text, followed by an MLP with gaussian error linear units (GELU, \cite{Hendrycks2016BridgingNA}) and two fully connected layers for classification. To preserve the general linguistic knowledge of the model while adapting it to the specific task at hand, we decided to ``freeze'' the RoBERTa model, allowing the loss to only backpropagate through the MLP module.

Let $E$ represent the input embedding, $V$ represent the vocabulary size, and $H$ denote the dimension of the last hidden state in RoBERTa. The input text with length $N$ can be expressed as a sequence of embedding, $E_{[CLS]},E_{1},\cdots,E_{N}$, where $E_{[CLS]},E_{i}\in \mathbb{R}^{V}$. Here, $E_{[CLS]}$ denotes the embedding of the special \texttt{[CLS]} token, as descried in the original BERT implementation \cite{BERT}. We use the final hidden state vector $T_{[CLS]} \in \mathbb{R}^{H}$ that corresponds to the first \texttt{[CLS]} token as the features of the input text. This extracted feature vector $C$ is then forward to the MLP for classification, as shown in figure \ref{fig:RoBERTa Architecture}.

The detailed training configuration for RoBERTa-Sentinel can be found in table \ref{tab: Training Configuration}.

\begin{table}
    \centering
    \caption{Training configuration for RoBERTa-Sentinel and T5-Sentinel. Where ``AdamW'' refers to the ``Adaptive Momentum Estimation with Weight Decay'' optimizer proposed by Loshchilov et al. in \cite{DBLP:journals/corr/abs-1711-05101}. ``Cosine annealing'' refers to the learning rate schedul proposed by Loschilov et al. in \cite{DBLP:journals/corr/LoshchilovH16a}}
    \begin{tabular}{lll}
    \toprule
    Hyper-Parameters & RoBERTa-Sentinel & T5-Sentinel \\
    \midrule
    Epoch           & $15$ & 5 \\
    Batch Size      & $512$ & $512$ \\
    Learning Rate   & $\num{1e-4}$ & $\num{5e-4}$\\
    Weight Decay    & $\num{1e-3}$ & \num{1e-3} \\
    Optimizer       & AdamW & AdamW\\
    Loss Function   & Cross entropy & Cross entropy\\
    Scheduler       & Cosine annealing & Cosine annealing  \\
    Data Set        & \texttt{OpenGPTText-Final} & \texttt{OpenGPTText-Final} \\
    \bottomrule
    \end{tabular}
    \label{tab: Training Configuration}
\end{table}

\subsection{T5-Sentinel Model}

The second methodwe proposed involves fine-tuning the T5 model \cite{t5-model} for classification tasks. Unlike the RoBERTa-Sentinel which uses an MLP module to classify the hidden state vector of input, this approach directly encodes the task as a sequence-to-sequence (seq-to-seq) problem.

\begin{figure}
    \centering
    \includegraphics[width=.8\textwidth]{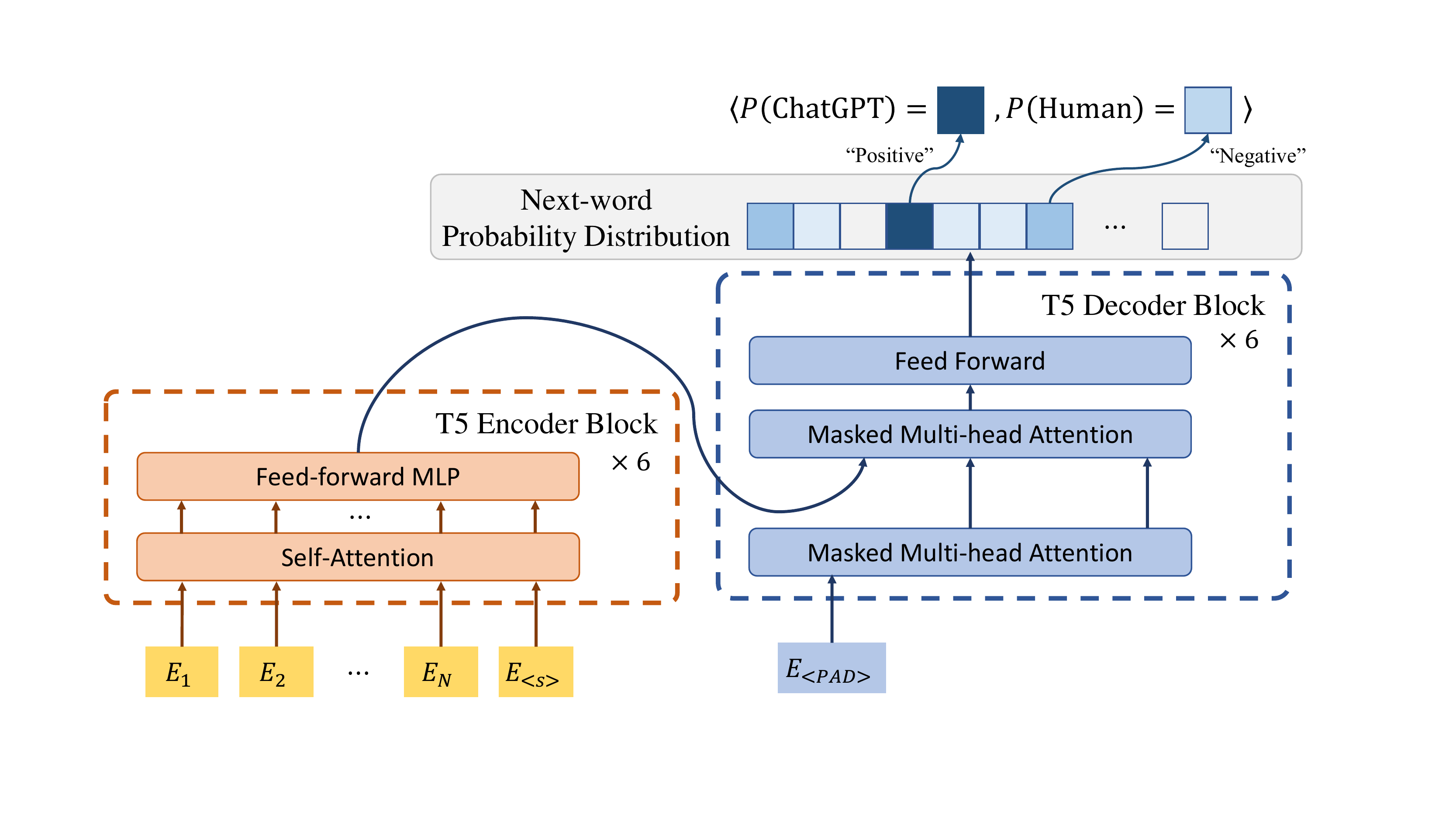}
    \caption{Architecture for T5-Sentinel. After input the entire token sequence, we provide the T5-Decoder with a \texttt{<PAD>} token and predict if the input text is by human or generated based on the probability of specific word ``Positive'' and ``Negative'' in the output word probability distribution.}
    \label{fig:T5 Architecture}
\end{figure}

During the training, the input sequence consists of a text sample from \texttt{OpenGPTText-Final}, and the output sequence represents the classification result as either ``positive </s>'' or ``negative </s>'', where ``</s>'' is the end-of-sequence token. During the inference, we limit the vocabulary down to only two words (i.e. ``positive'' and ``negative''), and select the one with the maximum probability as the classification result. This process is further shown in Figure \ref{fig:T5 Architecture}. 

The detailed training configuration of T5-Sentinel can be found in table \ref{tab: Training Configuration}.

\section{Evaluation}
\subsection{Evaluation Metric}

In our study, we assess the performance of the RoBERTa-Sentinel and T5-Sentinel through the application of five distinct evaluation metrics, namely the F1 score, receiver operating characteristic (ROC) curve, detection error trade-off (DET), area under curve (AUC), and model confidence score. In this paper, the term ``positive'' refers to the input text is ChatGPT-generated, while ``negative'' , means that the data is written by human.

Given the true positive ($TP$), true negative ($TN$), false positive ($FP$) and false negative ($FN$) count, we can calculate the metrics as following:


$$
\mathrm{F1\; Score} = \frac{TP}{TP + \frac{1}{2}(FP + FN)}
$$

$$
\mathrm{TPR} = \frac{TP}{TP + FN}\quad
\mathrm{FPR} = \frac{FP}{FP + TN}\quad
\mathrm{TNR} = \frac{TN}{TN + FP}\quad
\mathrm{FNR} = \frac{FN}{FN + TP}
$$

\subsection{F1 Score, False Positive Rate and False Negative Rate}

The F1 score, false positive rate and false negative rate of RoBERTa-Sentinel and T5-Sentinel is evaluated on original data set (\texttt{OpenGPTText}), cleaned data set (\texttt{OpenGPTText-Final}), and the \texttt{GPT2-Output}\footnote{There exist multiple variants of \texttt{GPT2-Output} data set, unless explicitly stated otherwise, in this paper we refer to the \texttt{GPT2-Output} data set with GPT2 Extra Large (1542M parameter) with pure sampling method.} data set \cite{GPT2-Output}. The evaluation results when taking $0.5$ as the threshold probability for positive are shown in the table \ref{tab:Accuracy Score}. For more detailed data on evaluation result, we included the true positive rate, true negative rate and sample count for each metric in table \ref{tab:Detailed Evaluation Data (Appendix)} in appendix B.

\begin{table}
    \caption{The evaluation result for T5-Sentinel, RoBERTa-Sentinel, ZeroGPT \cite{ZeroGPT}, OpenAI-Detector \cite{AITextClassifier}, and GPT-2 Detector from Solaiman et al. \cite{Solaiman2019} on three data sets under threshold probability of $0.5$. F1 stands for ``F1-score''. FPR and FNR data are in percentage.
    }
    \centering
    \begin{tabular}{lccccccccc}
    \toprule
    Model & \multicolumn{3}{c}{\texttt{OpenGPTText-Final}} & \multicolumn{3}{c}{\texttt{OpenGPTText}} & \multicolumn{3}{c}{GPT2-Output} \\
            \cmidrule{2-4}                          \cmidrule{5-7}                    \cmidrule{8-10}
          & F1 & FPR & FNR                       & F1 & FPR & FNR                 & F1 & FPR & FNR \\
    \midrule
    \textbf{T5}     & \textbf{0.98} & \textbf{2.8}  & \textbf{1.3}  & \textbf{0.98} & 3.5           & \textbf{1.3}  & 0.06          & \textbf{5.9}  & 96.7         \\
    RoBERTa         & 0.94          & 9.0           & 3.2           & 0.89          & 21.6          & 1.9           & 0.16          & 17.2          & 89.6         \\\midrule
    ZeroGPT         & 0.43          & 26.3          & 65.0          & 0.40          & 16.5          & 71.3          & 0.14          & 23.4          & 90.5         \\
    OpenAI-Detector & 0.32          & 4.9           & 79.8          & 0.26          & \textbf{1.6}  & 85.2          & 0.66          & 13.6          & 44.0         \\
    GPT2            & 0.23          & 2.8           & 86.8          & 0.22          & 4.1           & 87.2          & \textbf{0.93} & 6.4           & \textbf{7.4} \\
    \bottomrule
    \end{tabular}
    \label{tab:Accuracy Score}
\end{table}


An important observation is that even though T5-Sentinel and RoBERTa-Sentinel models exhibit high accuracy in the \texttt{OpenGPTText} data set, both prior to and post cleaning, they do not perform as effectively on the \texttt{GPT2-Output} data set, displaying an exceptionally high FNR. This disparity may be attributed to the distinctive quality of text generated by GPT2 and ChatGPT models, as well as the dissimilar nature of the samples in the \texttt{OpenGPTText} data set, which are all rephrased from human-written articles, in contrast to the \texttt{GPT2-Output} data set that contains randomly generated text.


Likewise, it is worth noting that the baseline model, GPT2-Detector by Solaiman et al., did not succeed in transferring its learned experience from the GPT2 output detection task to the task of detecting ChatGPT generated text, despite the findings presented in \cite{Solaiman2019}, which indicate that GPT2-Detector is capable of detecting diverse variants of the GPT2 model.

\subsection{ROC / DET Curve and AUC}

\begin{figure}
    \centering
    \begin{tabular}{ccc}
        \includegraphics[width=.3\textwidth]{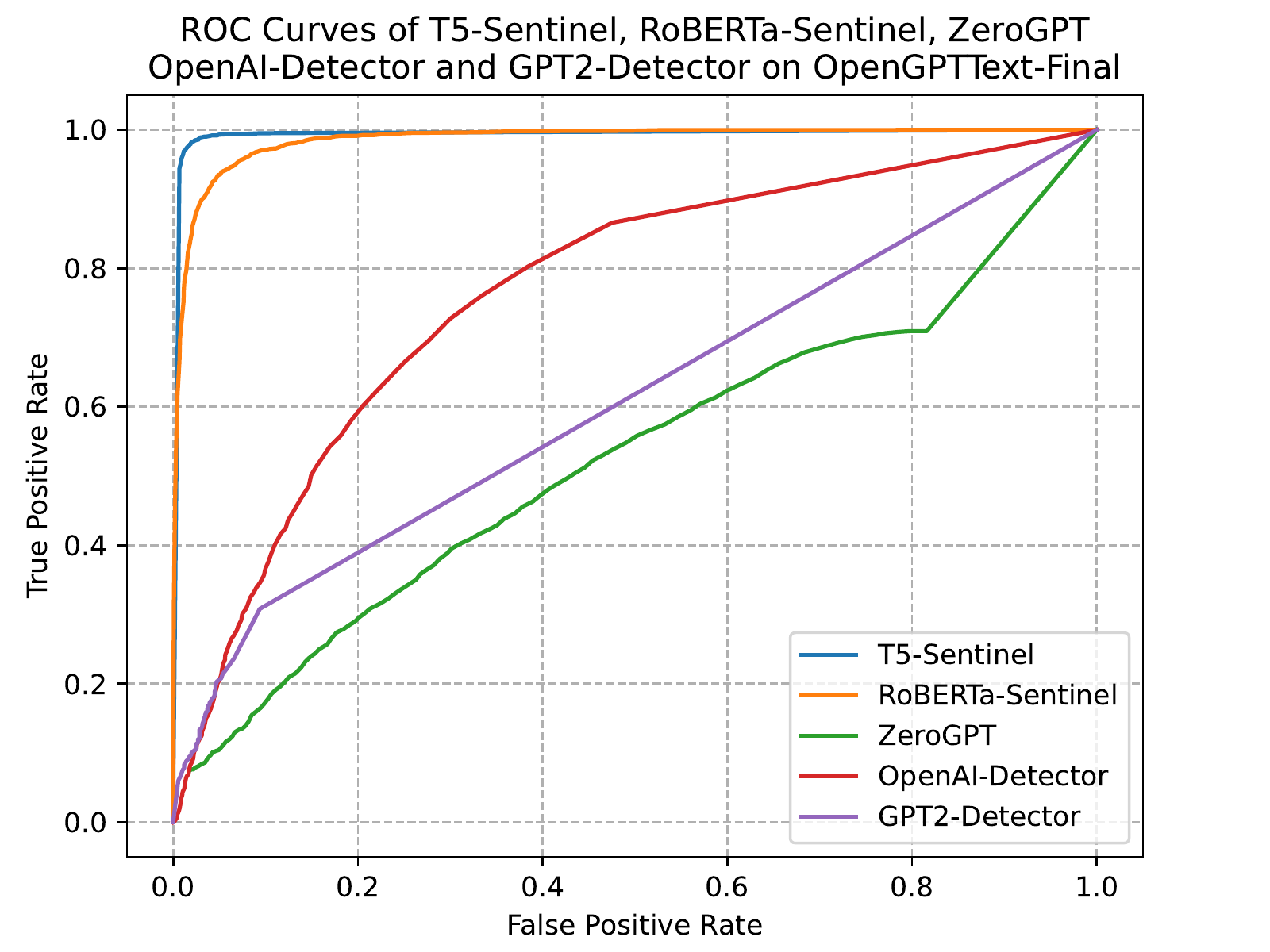} &
        \includegraphics[width=.3\textwidth]{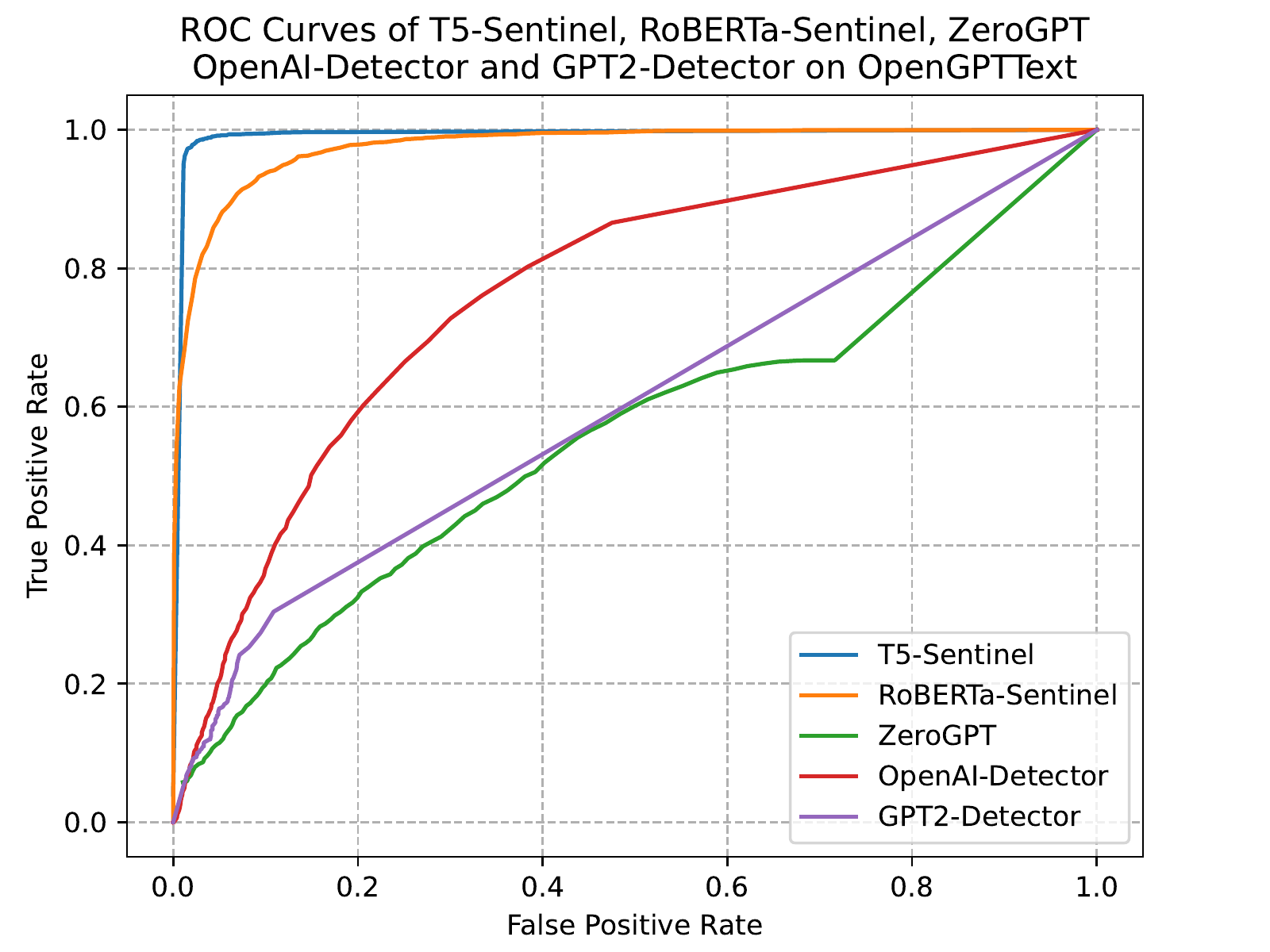} &
        \includegraphics[width=.3\textwidth]{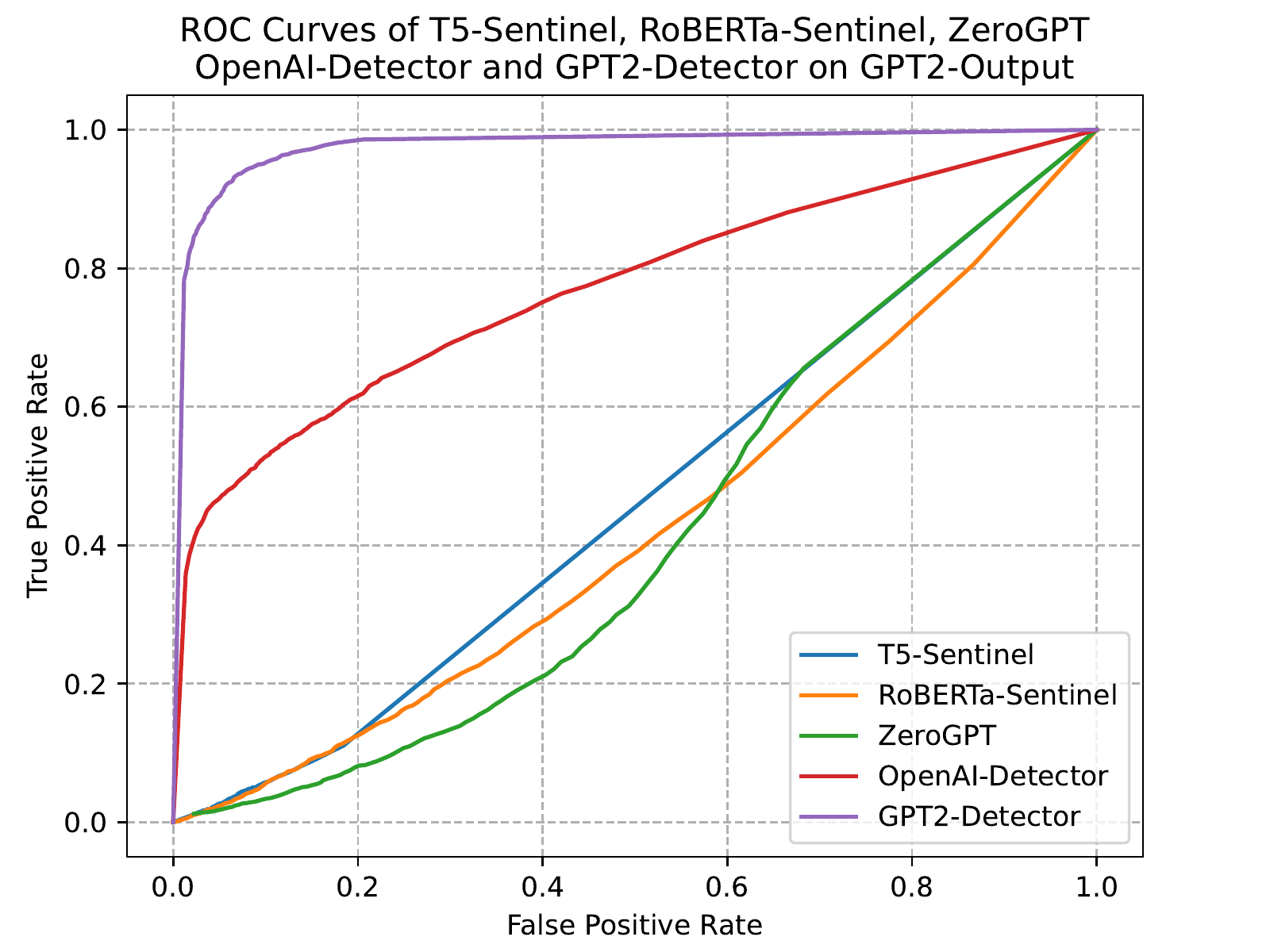}
    \end{tabular}
    \caption{ROC Curves for models across different data sets. \texttt{OpenGPTText-Final} (Left), \texttt{OpenGPTText} (Middle), and \texttt{GPT2-Output} (Right)}
    \label{fig:ROC Curves compare between models}
\end{figure}

\begin{figure}
    \centering
    \begin{tabular}{cc}
        \includegraphics[width=.4\textwidth]{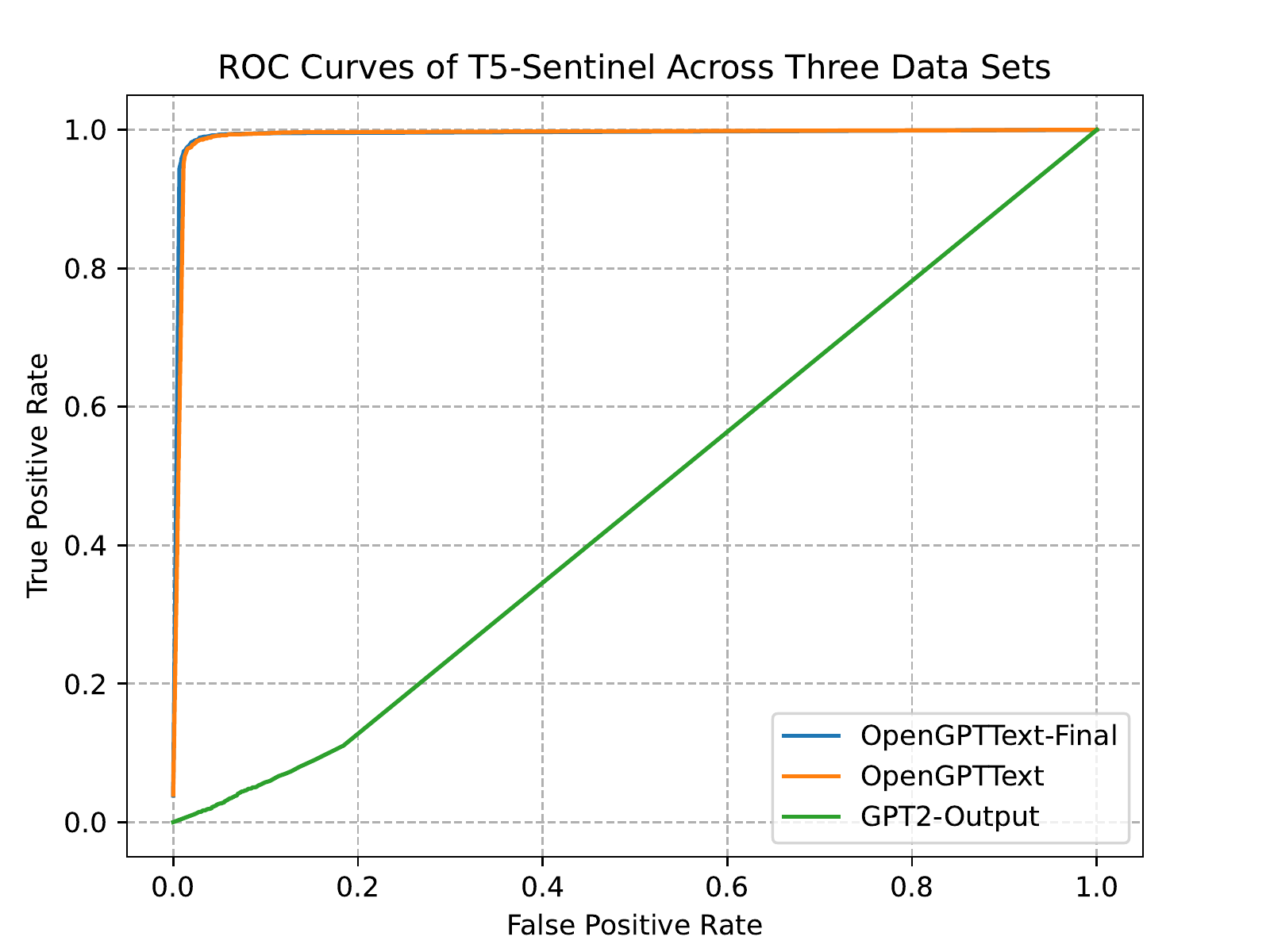} &
        \includegraphics[width=.4\textwidth]{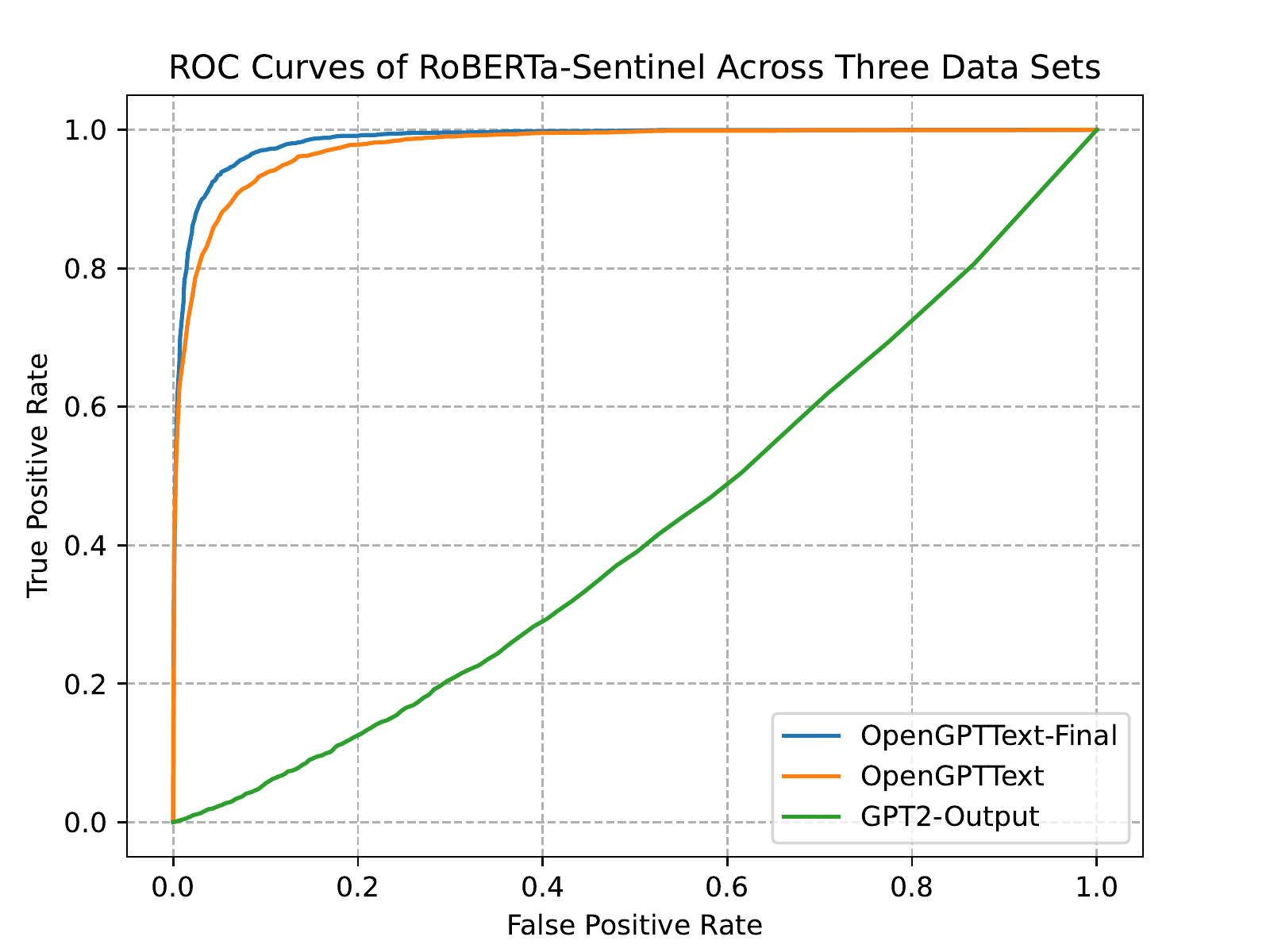}
    \end{tabular}
    \caption{ROC Curves for same model under different data sets T5-Sentinel (Left) and RoBERTa-Sentinel (Right). Note that the performance of RoBERTa-Sentinel significantly deteriorates when transfer to original version of \texttt{OpenGPTText} while T5-Sentinel does not.}
    \label{fig:ROC Curves compare between data sets}
\end{figure}

The ROC curve is a common graph used to evaluate and compare classifiers and can explicitly visualize the sensitivity/specificity trade-off of classifier for all thresholds \cite{Melo2013}. The ROC curve of T5-Sentinel, RoBERTa-Sentinel and GPT2-Detector on \texttt{OpenGPTText}, \texttt{OpenGPTText-Final} and \texttt{GPT2-Output} are shown in figure \ref{fig:ROC Curves compare between models} separately. Upon analyzing the ROC curves for the same model across different data sets, as illustrated in Figure \ref{fig:ROC Curves compare between data sets}, we observe that the T5-Sentinel  demonstrates greater robustness as compared to RoBERTa-Sentinel.

The area under curve (AUC) is a single-number summary for the ROC curve. The AUC result for each combination of model and dataset is listed in table \ref{tab:AUC for model and data sets}.

\begin{table}
    \caption{AUC Value for each combination of data set and model}
    \centering
    \begin{tabular}{lccc}
    \toprule
    Model & \texttt{OpenGPTText-Final} & \texttt{OpenGPTText} & \texttt{GPT2-Output} \\
    \midrule
    T5-Sentinel      & \textbf{0.993} & \textbf{0.992} & 0.463          \\
    RoBERTa-Sentinel & 0.986          & 0.976          & 0.423          \\ \midrule
    ZeroGPT          & 0.526          & 0.555          & 0.413          \\
    OpenAI-Detector  & 0.765          & 0.752          & 0.770          \\
    GPT2-Detector    & 0.610          & 0.600          & \textbf{0.976} \\
    \bottomrule
    \end{tabular}
    \label{tab:AUC for model and data sets}
\end{table}

\begin{figure}
    \centering
    \begin{tabular}{ccc}
        \includegraphics[width=.3\textwidth]{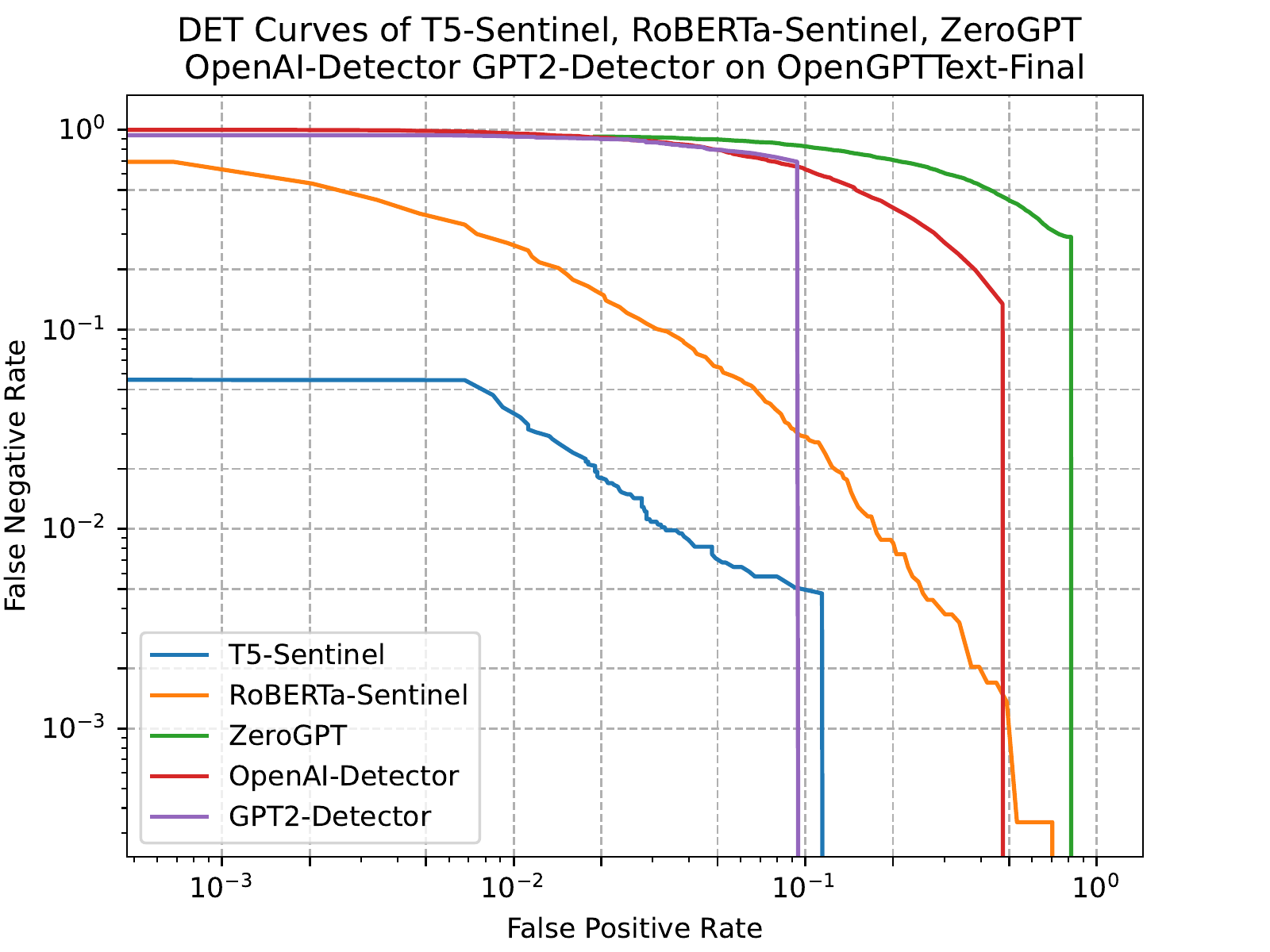} &
        \includegraphics[width=.3\textwidth]{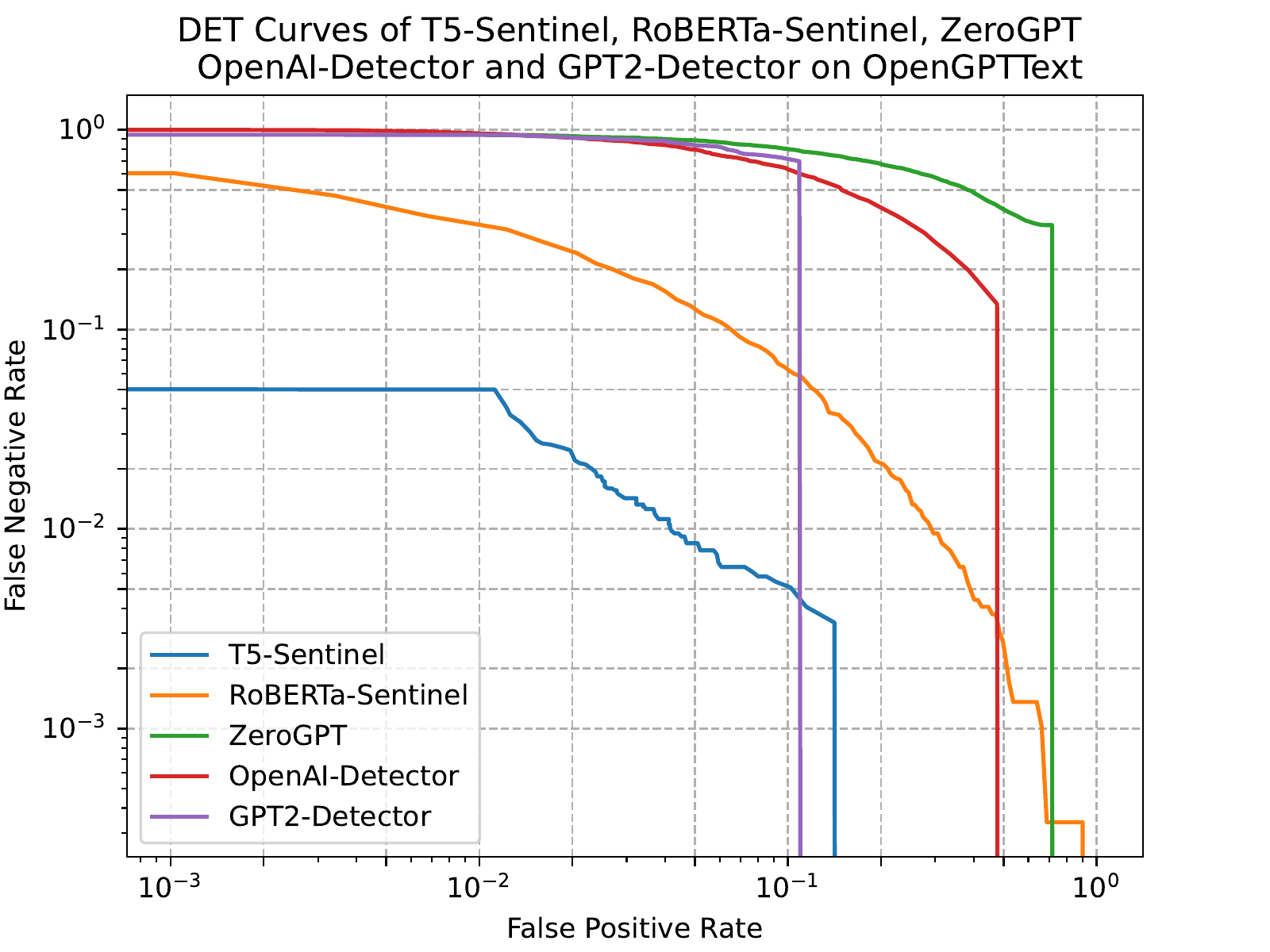} &
        \includegraphics[width=.3\textwidth]{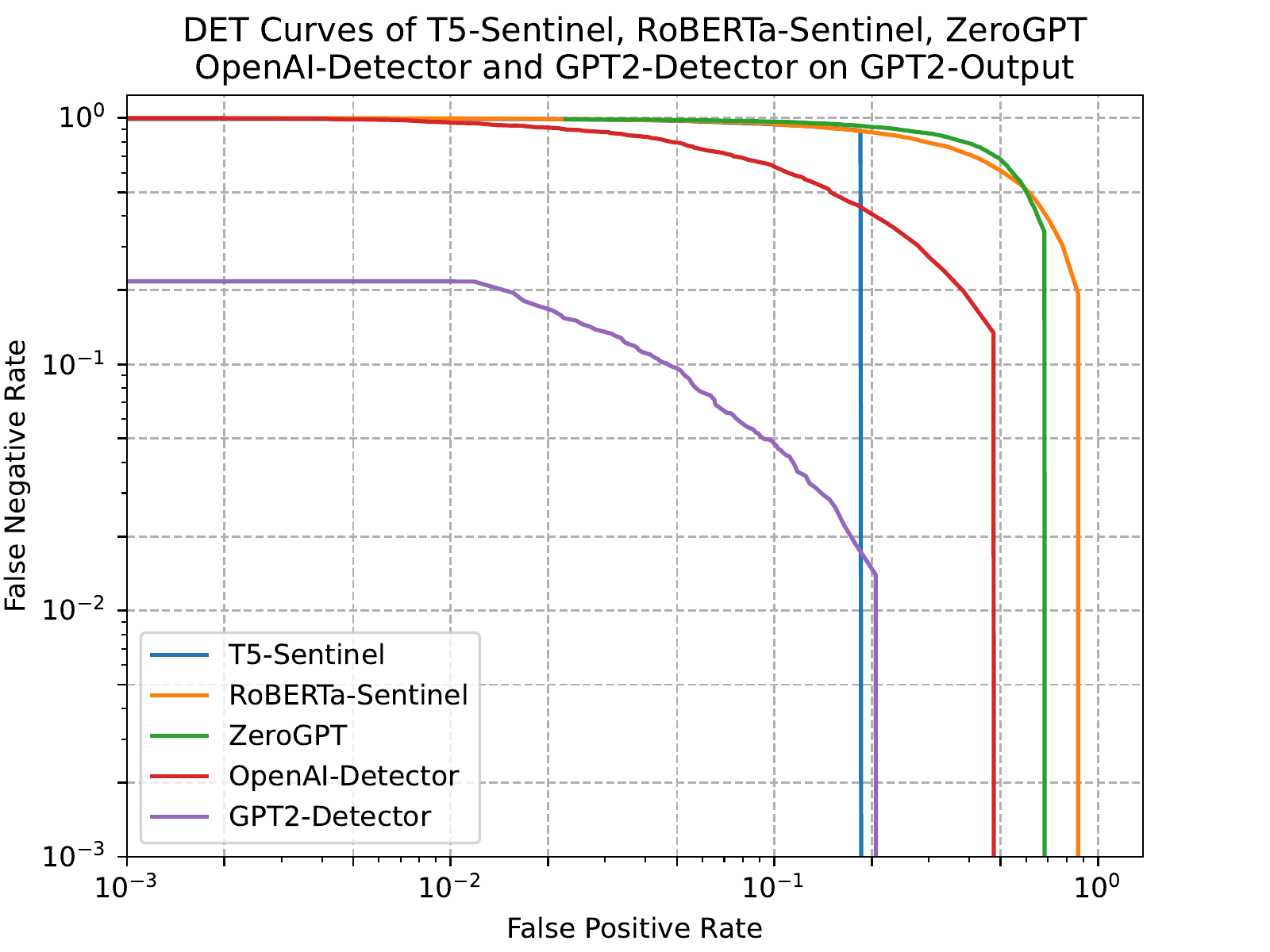}
    \end{tabular}
    \caption{DET Curves of different models under \texttt{OpenGPTText-Final} (Left), \texttt{OpenGPTText} (Middle) and \texttt{GPT2-Output} (Right) under logarithmic axis.}
    \label{fig:DET Curves compare between models}
\end{figure}

\begin{figure}
    \centering
    \begin{tabular}{cc}
        \includegraphics[width=.4\textwidth]{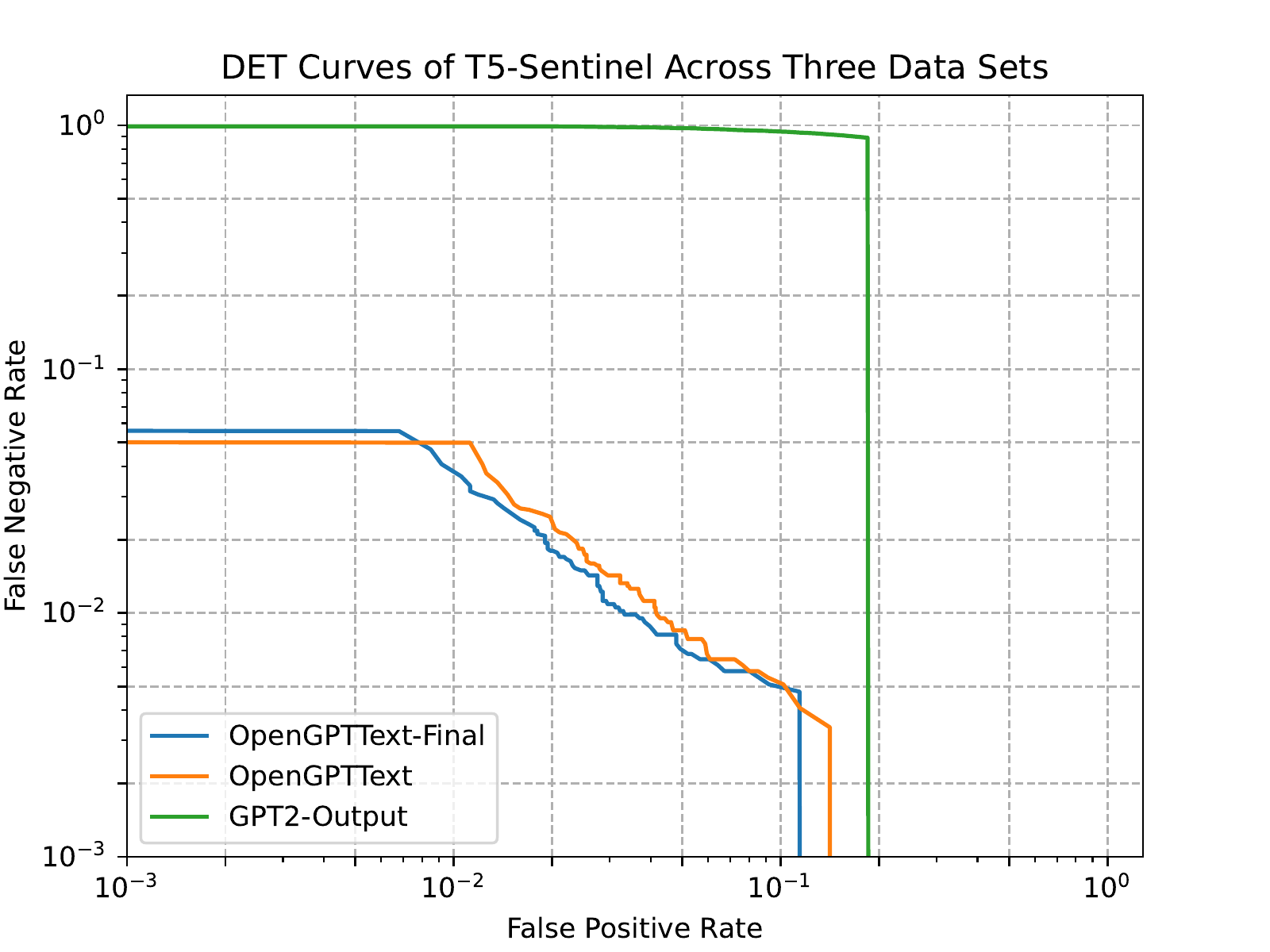} &
        \includegraphics[width=.4\textwidth]{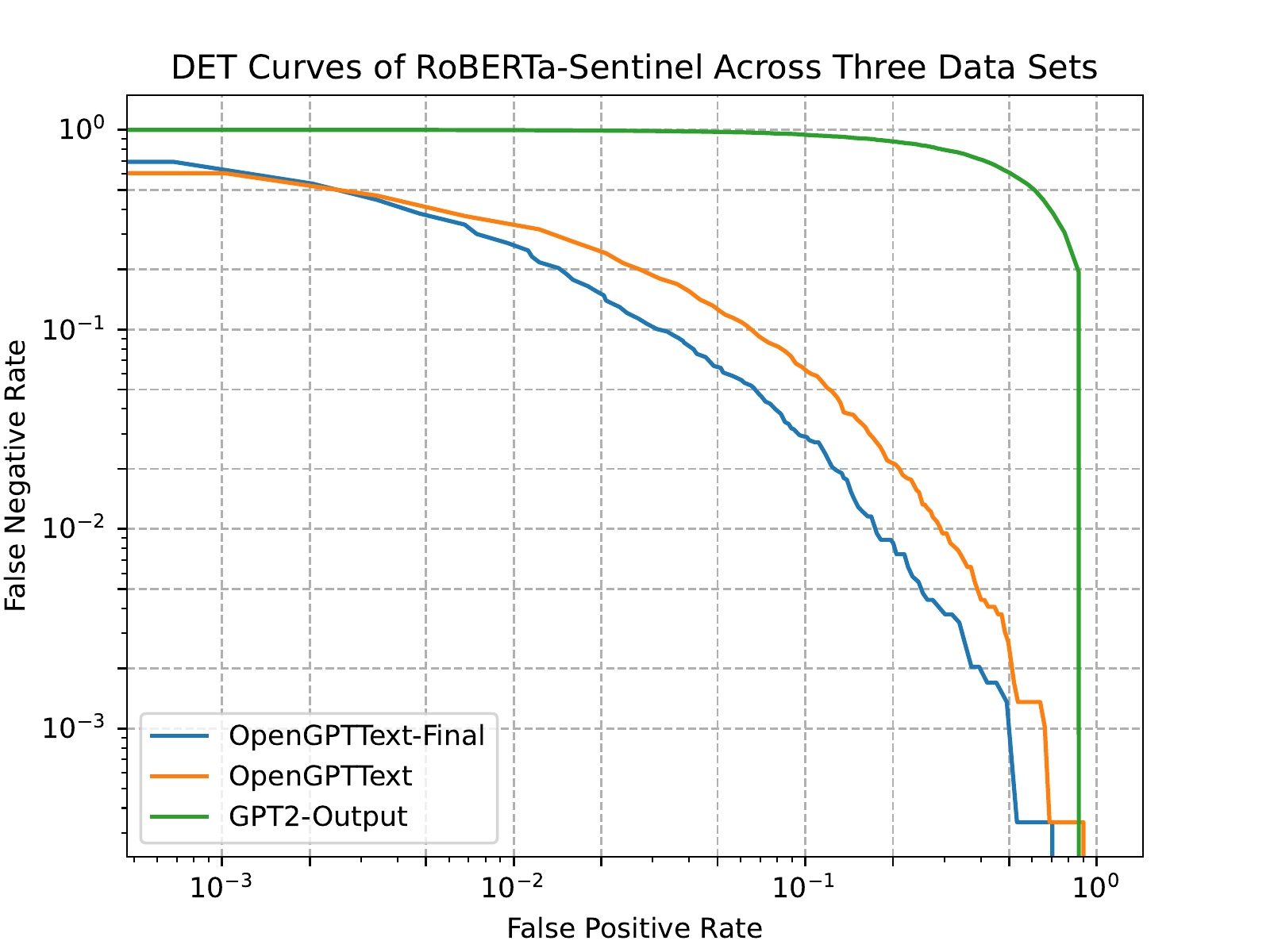}
    \end{tabular}
    \caption{DET Curves of T5-Sentinel (Left) and RoBERTa-Sentinel (Right) on different data sets.}
    \label{fig:DET Curves compare between data sets}
\end{figure}

We also plot the detector error trade-off (DET) curves across different models (figure \ref{fig:DET Curves compare between models}) and across different data sets (figure \ref{fig:DET Curves compare between data sets}).

\subsection{Confidence Score}
\begin{figure}
    \centering
    \begin{tabular}{ccc}
        \includegraphics[width=.45\textwidth]{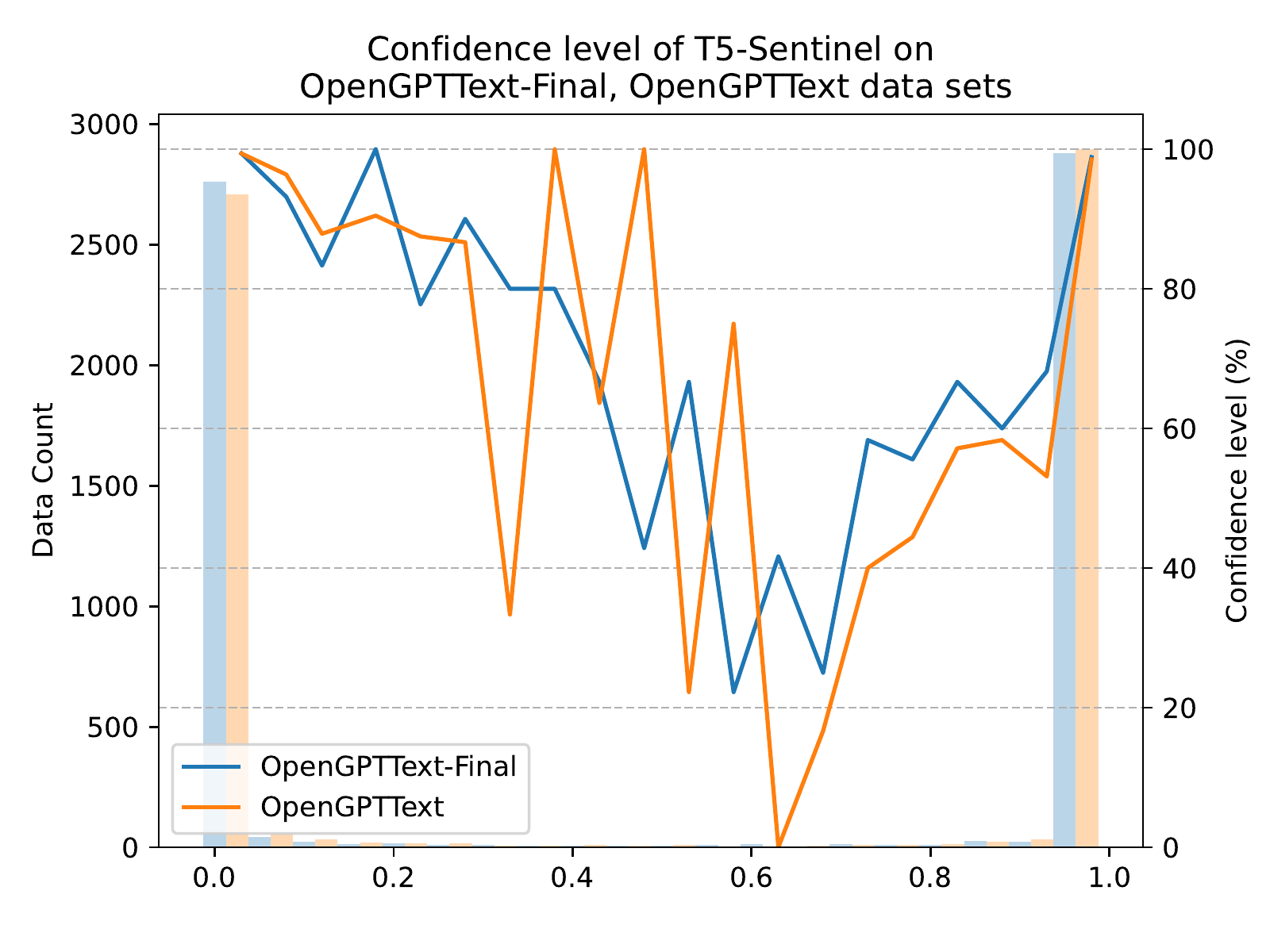} &
        \includegraphics[width=.45\textwidth]{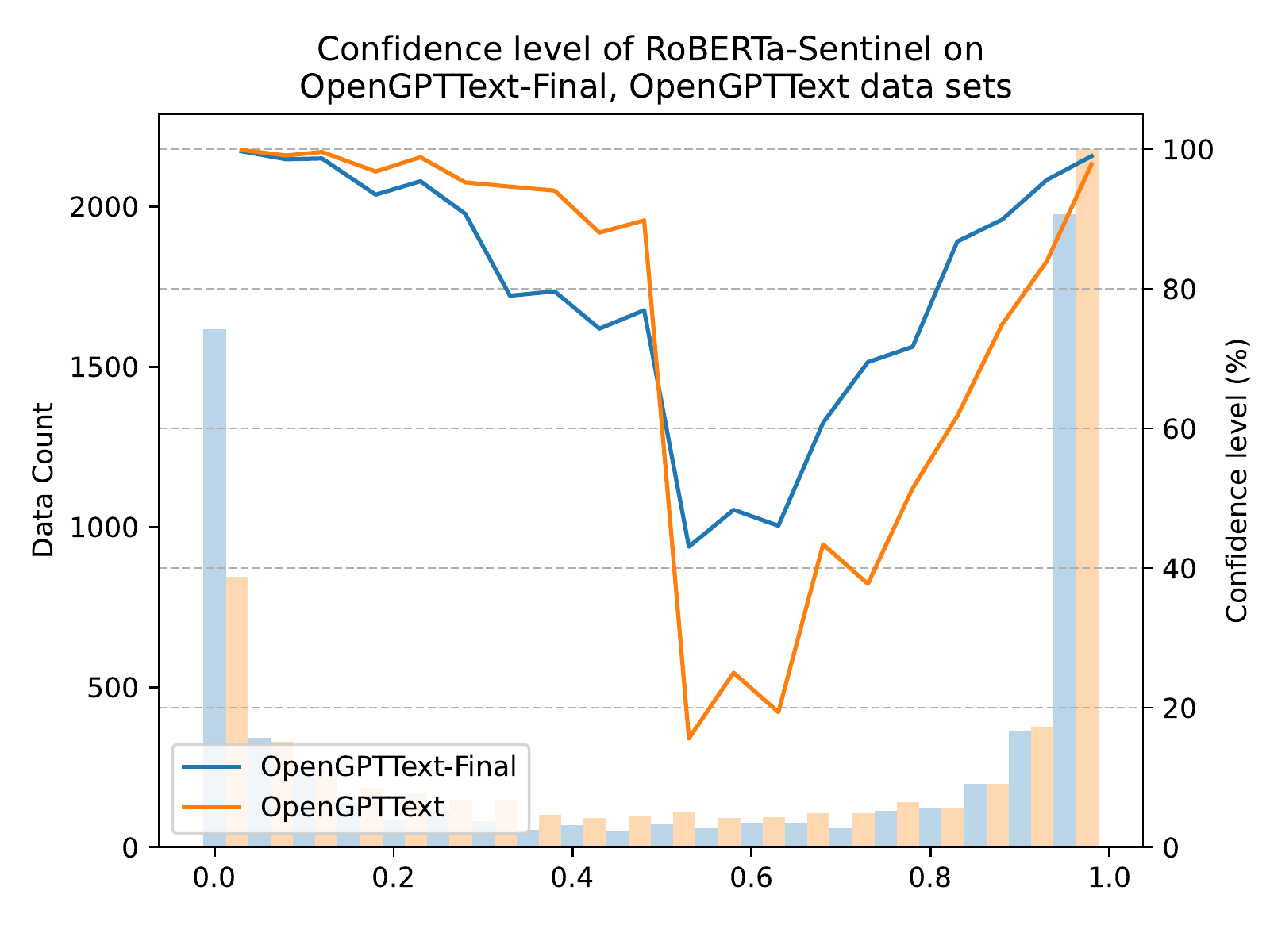}
    \end{tabular}
    \caption{Confidence Scores for T5-Sentinel and RoBERTa-Sentinel on \texttt{OpenGPTText-Final} and \texttt{OpenGPTText} data set. The histogram represents the number of sample under certain range of probability for the sample to be positive.}
    \label{fig:Confidence Score for each model}
\end{figure}

Assessing the reliability and confidence of a machine learning model's predictions is crucial for evaluating its performance. To this end, we calculated confidence scores for each combination of data sets and models, and plot them in figure \ref{fig:Confidence Score for each model}. The resulting confidence scores range from 0 to 1 and provide a measure of the model's certainty about its predictions.

In our analysis, we investigated the distribution of confidence scores and their correspondence with accuracy. Our result indicates that the T5-Sentinel model achieved higher confidence scores compared to RoBERTa-Sentinel. In contrast, the RoBERTa-Sentinel model had lower confidence scores than T5-Sentinel and showed greater confidence when detecting text generated by human than that by Chat-GPT.

Overall, our findings suggest that the T5-Sentinel model is more reliable and decisive than RoBERTa-Sentinel, particularly when dealing with \texttt{OpenGPTText}. Further investigation is needed to fully understand the reasons for these differences and to optimize the performance of these models for specific applications.

\section{Interpretability Study}

\subsection{Principal Component Analysis on Hidden State}

To offer greater insight into the functioning of the models we proposed, T5-Sentinel and RoBERTa-Sentinel, we conducted a PCA on their respective hidden states.

\begin{figure}
    \centering
    \begin{tabular}{ccc}
        \includegraphics[width=.45\textwidth]{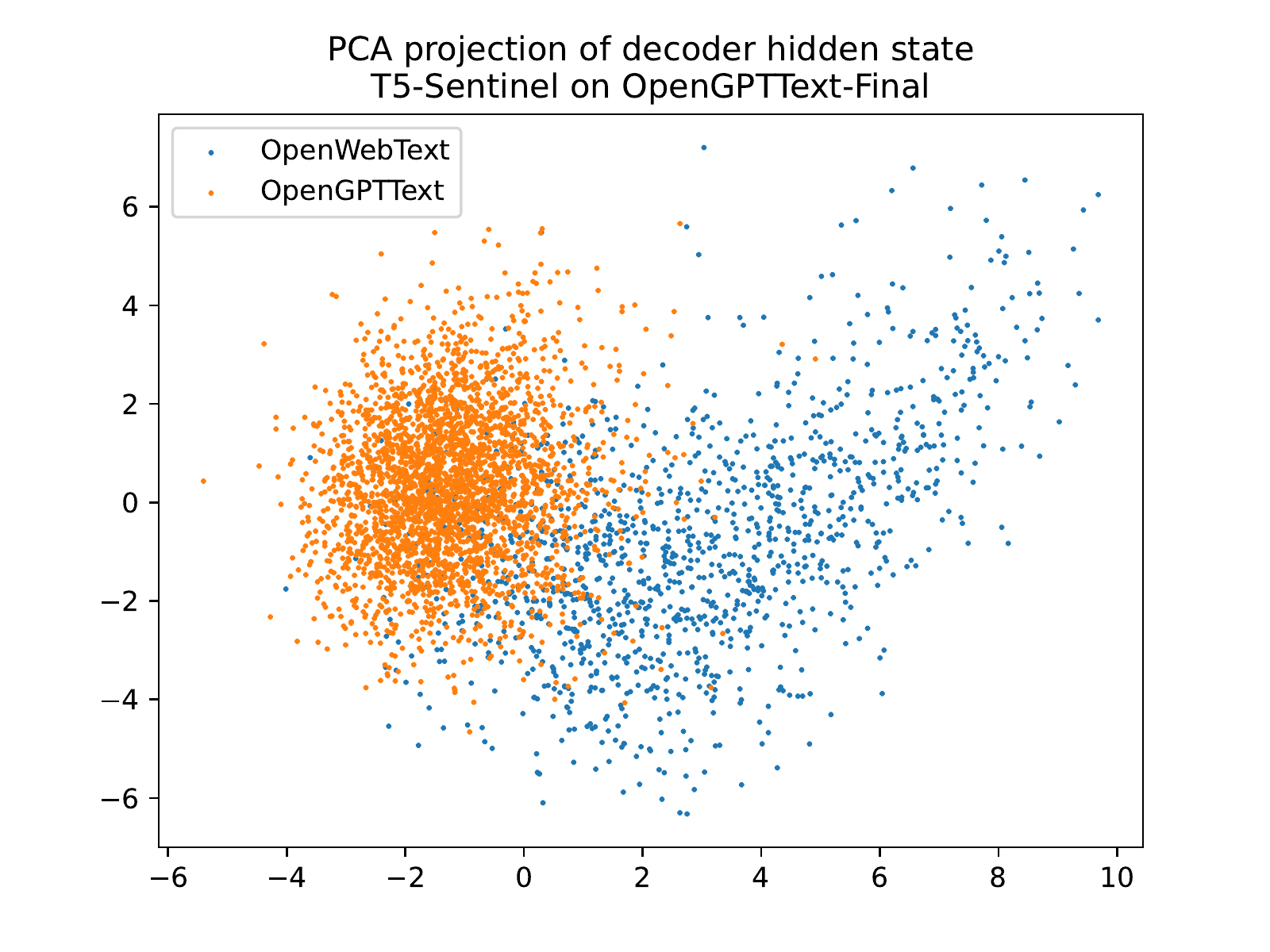} &
        \includegraphics[width=.45\textwidth]{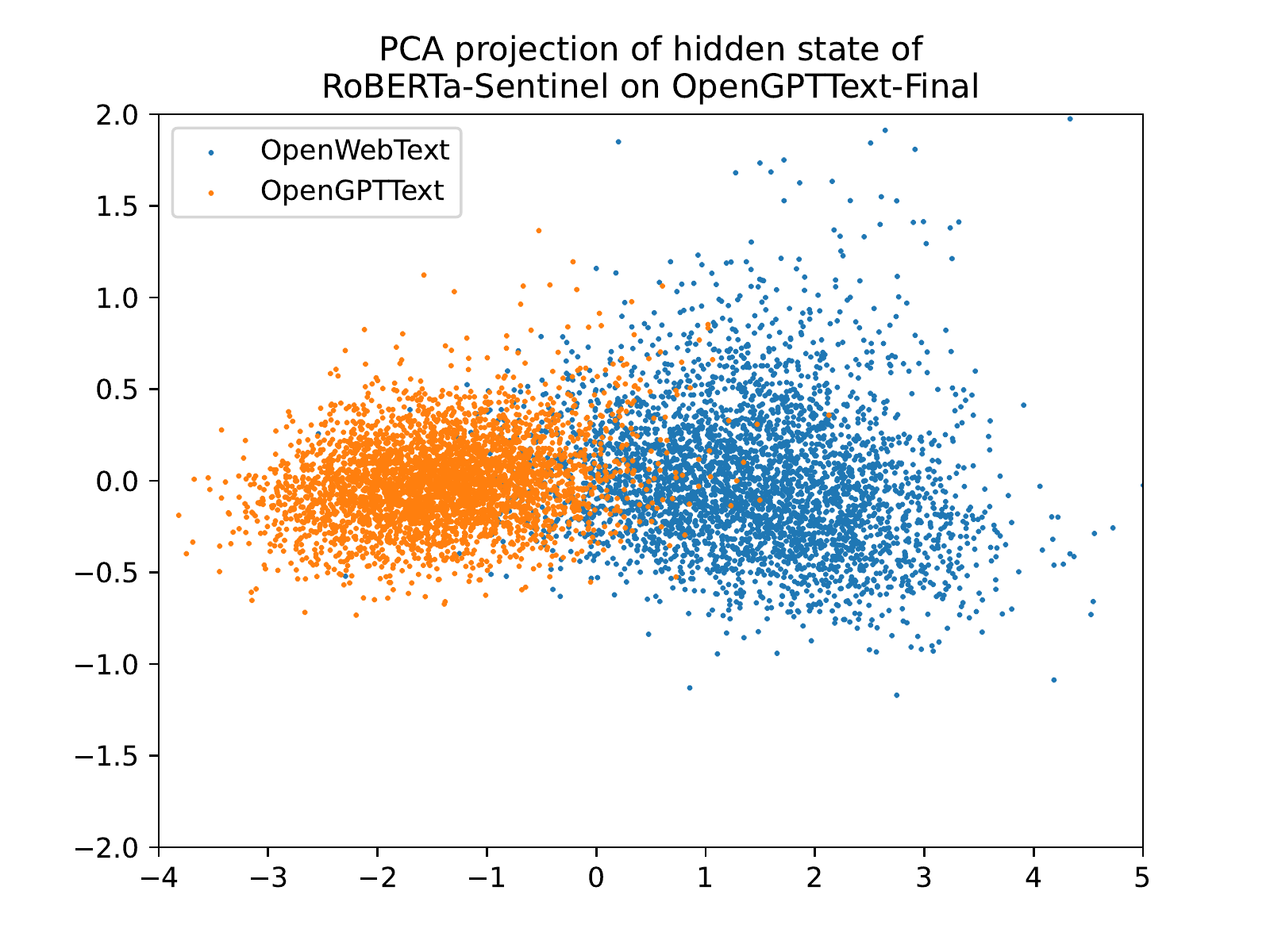}
    \end{tabular}
    \caption{PCA projection of hidden states for T5-Sentinel (Left) and RoBERTa-Sentinel (Right)}
    \label{fig:PCA Projection for Hidden States on OpenGPTText-test-final}
\end{figure}

For RoBERTa-Sentinel, we extracted the hidden state from last layer of attached MLP and the output of last decoder block for T5-Sentinel and recorded their values with input of all data in test set sampled from \texttt{OpenGPTText-Final}. As shown in figure \ref{fig:PCA Projection for Hidden States on OpenGPTText-test-final}, both models successfully mapped the input text into two different clusters in hidden space, indicating that both models were able to extract implicit characteristics of ChatGPT rephrased text.

To investigate the properties of the data along each direction of the projection subspace, we conducted a sampling of the data point outliers in the PCA projection subspace. 

\begin{figure}
    \centering
    \includegraphics[width=.6\textwidth]{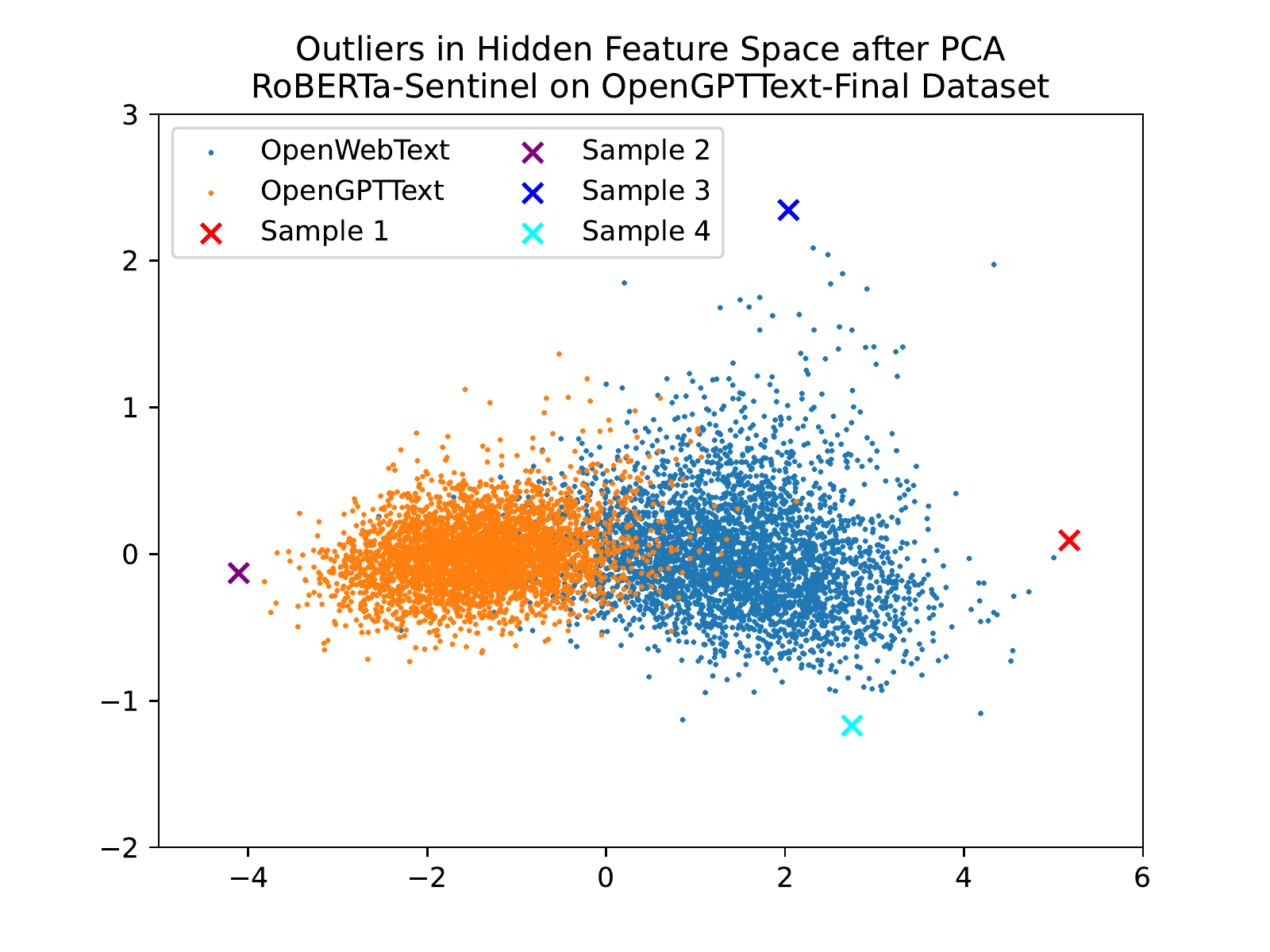}
    \caption{Outliers in PCA projection space drawn for manual inspection.}
    \label{fig:Outliers in PCA Projection Space}
\end{figure}

Figure \ref{fig:Outliers in PCA Projection Space} displays the position of four samples in the PCA projection subspace. Upon manual inspection of these samples, we discovered that \texttt{Sample 1} was a brief car advertisement that utilized simple language and comprised of very short paragraphs (split by images in the original web page). \texttt{Sample 2} was a sport news article with lengthy paragraphs, while \texttt{Sample 3} constituted a sequence of developing tool names that lacked any actual meaning. \texttt{Sample 4} was a brief report on children's attitudes towards clowns. These observations suggest that our model may have learned to distinguish the length of paragraphs and potentially discern whether a given text sample is informative and meaningful. We have provided detailed textual samples along with their unique identifiers (UIDs) from the \texttt{OpenGPTText} data set in Appendix B.

\subsection{Integrated Gradient}

We utilized the integrated gradient analysis technique, as proposed by Sundararajan et al. \cite{Sundararajan2017AxiomaticAF}, to gain insights into the contribution of individual tokens in a given input text towards the overall ``GPT-ness'' of the text. Our approach involved initially passing the input text through the model and computing the loss function under the assumption that the text label was ``human''. Following this, we executed back-propagation to obtain the gradients associated with each input token. This statement conformed to the formal style and technical language typically employed in academic writing.

The rationale underlying our method is rooted in the observation that tokens with gradients close to zero tend to align well with the human label, thereby necessitating minimal modification. Conversely, tokens with large gradients are indicative of a misalignment with the human label, suggesting a higher degree of resemblance to GPT-like characteristics.

We further developed a visualization tool that shows the contribution of each input token to the overall GPT-ness of the text. The darker the background token is, the more GPT-like that token is.

Below we show a sample\footnote{With data unique identifier (UID): [urlsf\_subset00]-[309279]} visualization result drawn from the test set of OpenGPTText-Final data set before and after the rephrasing. 

\textbf{Original Text:} \textit{Predict as human with probability of 0.998, with confidence of 0.994}

\begin{tcolorbox}[colback=white, colframe=black]
\ctext[RGB]{253,253,253}{Apple }\ctext[RGB]{235,235,235}{started }\ctext[RGB]{255,255,255}{an }\ctext[RGB]{255,255,255}{ }\ctext[RGB]{255,255,255}{ava }\ctext[RGB]{254,254,254}{lan }\ctext[RGB]{255,255,255}{che }\ctext[RGB]{255,255,255}{of }\ctext[RGB]{254,254,254}{activity }\ctext[RGB]{255,255,255}{with }\ctext[RGB]{247,247,247}{the }\ctext[RGB]{250,250,250}{introduction }\ctext[RGB]{255,255,255}{of }\ctext[RGB]{253,253,253}{the }\ctext[RGB]{255,255,255}{iPad }\ctext[RGB]{234,234,234}{. }\ctext[RGB]{239,239,239}{Companies }\ctext[RGB]{255,255,255}{ }\ctext[RGB]{253,253,253}{shifted }\ctext[RGB]{249,249,249}{gear }\ctext[RGB]{247,247,247}{s }\ctext[RGB]{248,248,248}{to }\ctext[RGB]{255,255,255}{go }\ctext[RGB]{247,247,247}{after }\ctext[RGB]{245,245,245}{this }\ctext[RGB]{254,254,254}{un }\ctext[RGB]{232,232,232}{discovered }\ctext[RGB]{255,255,255}{new }\ctext[RGB]{255,255,255}{tablet }\ctext[RGB]{254,254,254}{market }\ctext[RGB]{242,242,242}{. }\ctext[RGB]{236,236,236}{In }\ctext[RGB]{229,229,229}{spite }\ctext[RGB]{249,249,249}{of }\ctext[RGB]{255,255,255}{the }\ctext[RGB]{255,255,255}{number }\ctext[RGB]{255,255,255}{of }\ctext[RGB]{244,244,244}{players }\ctext[RGB]{245,245,245}{in }\ctext[RGB]{255,255,255}{tablets }\ctext[RGB]{255,255,255}{, }\ctext[RGB]{245,245,245}{no }\ctext[RGB]{247,247,247}{company }\ctext[RGB]{246,246,246}{has }\ctext[RGB]{225,225,225}{discovered }\ctext[RGB]{255,255,255}{the }\ctext[RGB]{242,242,242}{magic }\ctext[RGB]{253,253,253}{bullet }\ctext[RGB]{252,252,252}{to }\ctext[RGB]{248,248,248}{knock }\ctext[RGB]{254,254,254}{the }\ctext[RGB]{243,243,243}{iPad }\ctext[RGB]{249,249,249}{off }\ctext[RGB]{253,253,253}{the }\ctext[RGB]{243,243,243}{top }\ctext[RGB]{251,251,251}{of }\ctext[RGB]{255,255,255}{the }\ctext[RGB]{240,240,240}{tablet }\ctext[RGB]{255,255,255}{heap }\ctext[RGB]{222,222,222}{. }\ctext[RGB]{223,223,223}{Col }\ctext[RGB]{237,237,237}{league }\ctext[RGB]{238,238,238}{Adrian }\ctext[RGB]{255,255,255}{Kings }\ctext[RGB]{255,255,255}{ley }\ctext[RGB]{249,249,249}{- }\ctext[RGB]{255,255,255}{H }\ctext[RGB]{255,255,255}{ugh }\ctext[RGB]{255,255,255}{e }\ctext[RGB]{239,239,239}{s }\ctext[RGB]{227,227,227}{has }\ctext[RGB]{245,245,245}{ }\ctext[RGB]{249,249,249}{a }\ctext[RGB]{225,225,225}{thoughtful }\ctext[RGB]{226,226,226}{piece }\ctext[RGB]{251,251,251}{ }\ctext[RGB]{226,226,226}{bla }\ctext[RGB]{244,244,244}{m }\ctext[RGB]{247,247,247}{ing }\ctext[RGB]{255,255,255}{Amazon }\ctext[RGB]{245,245,245}{and }\ctext[RGB]{255,255,255}{Google }\ctext[RGB]{240,240,240}{for }\ctext[RGB]{238,238,238}{killing }\ctext[RGB]{249,249,249}{the }\ctext[RGB]{248,248,248}{tablet }\ctext[RGB]{245,245,245}{market }\ctext[RGB]{233,233,233}{. }\ctext[RGB]{250,250,250}{His }\ctext[RGB]{244,244,244}{reasoning }\ctext[RGB]{237,237,237}{is }\ctext[RGB]{244,244,244}{that }\ctext[RGB]{253,253,253}{by }\ctext[RGB]{254,254,254}{ }\ctext[RGB]{255,255,255}{releasing }\ctext[RGB]{250,250,250}{the }\ctext[RGB]{253,253,253}{ }\ctext[RGB]{255,255,255}{Kindle }\ctext[RGB]{254,254,254}{Fire }\ctext[RGB]{255,255,255}{and }\ctext[RGB]{255,255,255}{the }\ctext[RGB]{251,251,251}{Nex }\ctext[RGB]{255,255,255}{us }\ctext[RGB]{255,255,255}{7 }\ctext[RGB]{255,255,255}{at }\ctext[RGB]{255,255,255}{\$1 }\ctext[RGB]{255,255,255}{99 }\ctext[RGB]{243,243,243}{, }\ctext[RGB]{246,246,246}{Amazon }\ctext[RGB]{241,241,241}{and }\ctext[RGB]{255,255,255}{Google }\ctext[RGB]{255,255,255}{have }\ctext[RGB]{249,249,249}{started }\ctext[RGB]{252,252,252}{ }\ctext[RGB]{244,244,244}{a }\ctext[RGB]{255,255,255}{" }\ctext[RGB]{249,249,249}{race }\ctext[RGB]{253,253,253}{to }\ctext[RGB]{255,255,255}{the }\ctext[RGB]{255,255,255}{bottom }\ctext[RGB]{241,241,241}{" }\ctext[RGB]{254,254,254}{of }\ctext[RGB]{255,255,255}{the }\ctext[RGB]{253,253,253}{tablet }\ctext[RGB]{251,251,251}{market }\ctext[RGB]{240,240,240}{that }\ctext[RGB]{251,251,251}{will }\ctext[RGB]{238,238,238}{ensure }\ctext[RGB]{254,254,254}{no }\ctext[RGB]{253,253,253}{profitability }\ctext[RGB]{254,254,254}{for }\ctext[RGB]{252,252,252}{anyone }\ctext[RGB]{234,234,234}{. }\ctext[RGB]{234,234,234}{Adrian }\ctext[RGB]{255,255,255}{' }\ctext[RGB]{247,247,247}{s }\ctext[RGB]{239,239,239}{reasoning }\ctext[RGB]{254,254,254}{is }\ctext[RGB]{231,231,231}{solid }\ctext[RGB]{237,237,237}{, }\ctext[RGB]{239,239,239}{but }\ctext[RGB]{244,244,244}{it }\ctext[RGB]{245,245,245}{overlook }\ctext[RGB]{245,245,245}{s }\ctext[RGB]{236,236,236}{one }\ctext[RGB]{235,235,235}{thing }\ctext[RGB]{214,214,214}{I }\ctext[RGB]{230,230,230}{have }\ctext[RGB]{242,242,242}{said }\ctext[RGB]{241,241,241}{for }\ctext[RGB]{246,246,246}{ }\ctext[RGB]{242,242,242}{a }\ctext[RGB]{255,255,255}{long }\ctext[RGB]{241,241,241}{time }\ctext[RGB]{245,245,245}{. }\ctext[RGB]{251,251,251}{There }\ctext[RGB]{235,235,235}{is }\ctext[RGB]{248,248,248}{no }\ctext[RGB]{248,248,248}{proven }\ctext[RGB]{255,255,255}{tablet }\ctext[RGB]{255,255,255}{market }\ctext[RGB]{240,240,240}{. } (... Truncated)
\end{tcolorbox}

\textbf{Rephrased Text:} \textit{Predict as generated with probability of 1.000, with confidence of 0.985}

\begin{tcolorbox}[colback=white, colframe=black]
\ctext[RGB]{179,179,179}{Following }\ctext[RGB]{195,195,195}{the }\ctext[RGB]{215,215,215}{release }\ctext[RGB]{196,196,196}{of }\ctext[RGB]{195,195,195}{the }\ctext[RGB]{213,213,213}{iPad }\ctext[RGB]{180,180,180}{, }\ctext[RGB]{185,185,185}{the }\ctext[RGB]{212,212,212}{tablet }\ctext[RGB]{208,208,208}{market }\ctext[RGB]{181,181,181}{became }\ctext[RGB]{192,192,192}{ }\ctext[RGB]{213,213,213}{a }\ctext[RGB]{189,189,189}{popular }\ctext[RGB]{200,200,200}{area }\ctext[RGB]{194,194,194}{for }\ctext[RGB]{202,202,202}{companies }\ctext[RGB]{191,191,191}{to }\ctext[RGB]{179,179,179}{explore }\ctext[RGB]{181,181,181}{. }\ctext[RGB]{202,202,202}{ }\ctext[RGB]{177,177,177}{Despite }\ctext[RGB]{194,194,194}{many }\ctext[RGB]{185,185,185}{companies }\ctext[RGB]{185,185,185}{entering }\ctext[RGB]{198,198,198}{the }\ctext[RGB]{188,188,188}{market }\ctext[RGB]{201,201,201}{, }\ctext[RGB]{189,189,189}{no }\ctext[RGB]{189,189,189}{one }\ctext[RGB]{188,188,188}{has }\ctext[RGB]{212,212,212}{managed }\ctext[RGB]{200,200,200}{to }\ctext[RGB]{193,193,193}{out }\ctext[RGB]{228,228,228}{per }\ctext[RGB]{190,190,190}{form }\ctext[RGB]{216,216,216}{the }\ctext[RGB]{198,198,198}{iPad }\ctext[RGB]{206,206,206}{on }\ctext[RGB]{195,195,195}{sales }\ctext[RGB]{178,178,178}{. }\ctext[RGB]{177,177,177}{There }\ctext[RGB]{185,185,185}{is }\ctext[RGB]{193,193,193}{ }\ctext[RGB]{207,207,207}{a }\ctext[RGB]{200,200,200}{belief }\ctext[RGB]{183,183,183}{that }\ctext[RGB]{199,199,199}{Amazon }\ctext[RGB]{186,186,186}{and }\ctext[RGB]{208,208,208}{Google }\ctext[RGB]{188,188,188}{have }\ctext[RGB]{194,194,194}{caused }\ctext[RGB]{182,182,182}{this }\ctext[RGB]{172,172,172}{, }\ctext[RGB]{182,182,182}{due }\ctext[RGB]{189,189,189}{to }\ctext[RGB]{182,182,182}{the }\ctext[RGB]{190,190,190}{release }\ctext[RGB]{180,180,180}{of }\ctext[RGB]{181,181,181}{the }\ctext[RGB]{194,194,194}{ }\ctext[RGB]{198,198,198}{Kindle }\ctext[RGB]{196,196,196}{Fire }\ctext[RGB]{202,202,202}{and }\ctext[RGB]{209,209,209}{Nex }\ctext[RGB]{204,204,204}{us }\ctext[RGB]{200,200,200}{7 }\ctext[RGB]{203,203,203}{at }\ctext[RGB]{189,189,189}{\$1 }\ctext[RGB]{209,209,209}{99 }\ctext[RGB]{169,169,169}{, }\ctext[RGB]{197,197,197}{starting }\ctext[RGB]{201,201,201}{ }\ctext[RGB]{190,190,190}{a }\ctext[RGB]{181,181,181}{" }\ctext[RGB]{192,192,192}{race }\ctext[RGB]{193,193,193}{to }\ctext[RGB]{216,216,216}{the }\ctext[RGB]{207,207,207}{bottom }\ctext[RGB]{181,181,181}{." }\ctext[RGB]{179,179,179}{There }\ctext[RGB]{200,200,200}{is }\ctext[RGB]{209,209,209}{ }\ctext[RGB]{202,202,202}{a }\ctext[RGB]{181,181,181}{concern }\ctext[RGB]{195,195,195}{that }\ctext[RGB]{207,207,207}{this }\ctext[RGB]{192,192,192}{will }\ctext[RGB]{186,186,186}{ensure }\ctext[RGB]{197,197,197}{there }\ctext[RGB]{193,193,193}{will }\ctext[RGB]{198,198,198}{be }\ctext[RGB]{187,187,187}{no }\ctext[RGB]{203,203,203}{profitability }\ctext[RGB]{201,201,201}{for }\ctext[RGB]{209,209,209}{anyone }\ctext[RGB]{186,186,186}{. }\ctext[RGB]{170,170,170}{However }\ctext[RGB]{181,181,181}{, }\ctext[RGB]{182,182,182}{this }\ctext[RGB]{187,187,187}{belief }\ctext[RGB]{180,180,180}{overlook }\ctext[RGB]{189,189,189}{s }\ctext[RGB]{195,195,195}{the }\ctext[RGB]{185,185,185}{fact }\ctext[RGB]{187,187,187}{that }\ctext[RGB]{188,188,188}{there }\ctext[RGB]{202,202,202}{is }\ctext[RGB]{197,197,197}{no }\ctext[RGB]{208,208,208}{proven }\ctext[RGB]{216,216,216}{market }\ctext[RGB]{184,184,184}{for }\ctext[RGB]{192,192,192}{tablets }\ctext[RGB]{173,173,173}{, }\ctext[RGB]{185,185,185}{only }\ctext[RGB]{198,198,198}{ }\ctext[RGB]{194,194,194}{a }\ctext[RGB]{195,195,195}{proven }\ctext[RGB]{200,200,200}{market }\ctext[RGB]{187,187,187}{for }\ctext[RGB]{207,207,207}{iPad }\ctext[RGB]{217,217,217}{s }\ctext[RGB]{182,182,182}{. }\ctext[RGB]{178,178,178}{Other }\ctext[RGB]{180,180,180}{than }\ctext[RGB]{187,187,187}{Apple }\ctext[RGB]{187,187,187}{, }\ctext[RGB]{190,190,190}{Samsung }\ctext[RGB]{198,198,198}{is }\ctext[RGB]{190,190,190}{the }\ctext[RGB]{197,197,197}{only }\ctext[RGB]{201,201,201}{company }\ctext[RGB]{194,194,194}{with }\ctext[RGB]{191,191,191}{notable }\ctext[RGB]{196,196,196}{tablet }\ctext[RGB]{190,190,190}{sales }\ctext[RGB]{189,189,189}{. }\ctext[RGB]{187,187,187}{Even }\ctext[RGB]{185,185,185}{though }\ctext[RGB]{190,190,190}{they }\ctext[RGB]{201,201,201}{released }\ctext[RGB]{196,196,196}{several }\ctext[RGB]{208,208,208}{tablets }\ctext[RGB]{180,180,180}{, }\ctext[RGB]{211,211,211}{in }\ctext[RGB]{188,188,188}{various }\ctext[RGB]{196,196,196}{shapes }\ctext[RGB]{201,201,201}{and }\ctext[RGB]{201,201,201}{sizes }\ctext[RGB]{190,190,190}{, }\ctext[RGB]{201,201,201}{they }\ctext[RGB]{181,181,181}{still }\ctext[RGB]{196,196,196}{could }\ctext[RGB]{216,216,216}{not }\ctext[RGB]{197,197,197}{compete }\ctext[RGB]{191,191,191}{with }\ctext[RGB]{209,209,209}{the }\ctext[RGB]{215,215,215}{iPad }\ctext[RGB]{180,180,180}{. } (...Truncated)
\end{tcolorbox}

\begin{figure}
    \centering
    \includegraphics[width=.45\textwidth]{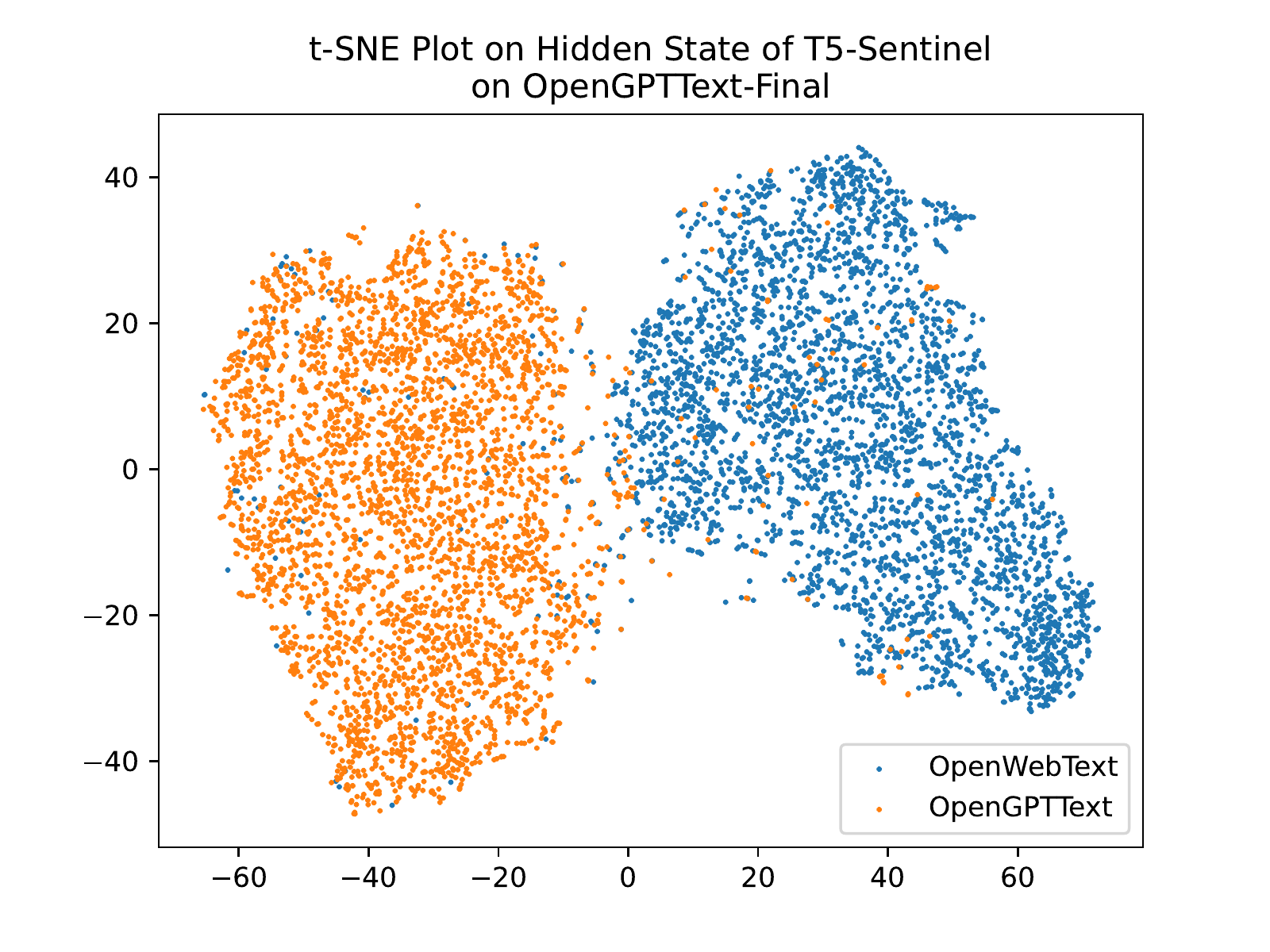}
    \includegraphics[width=.45\textwidth]{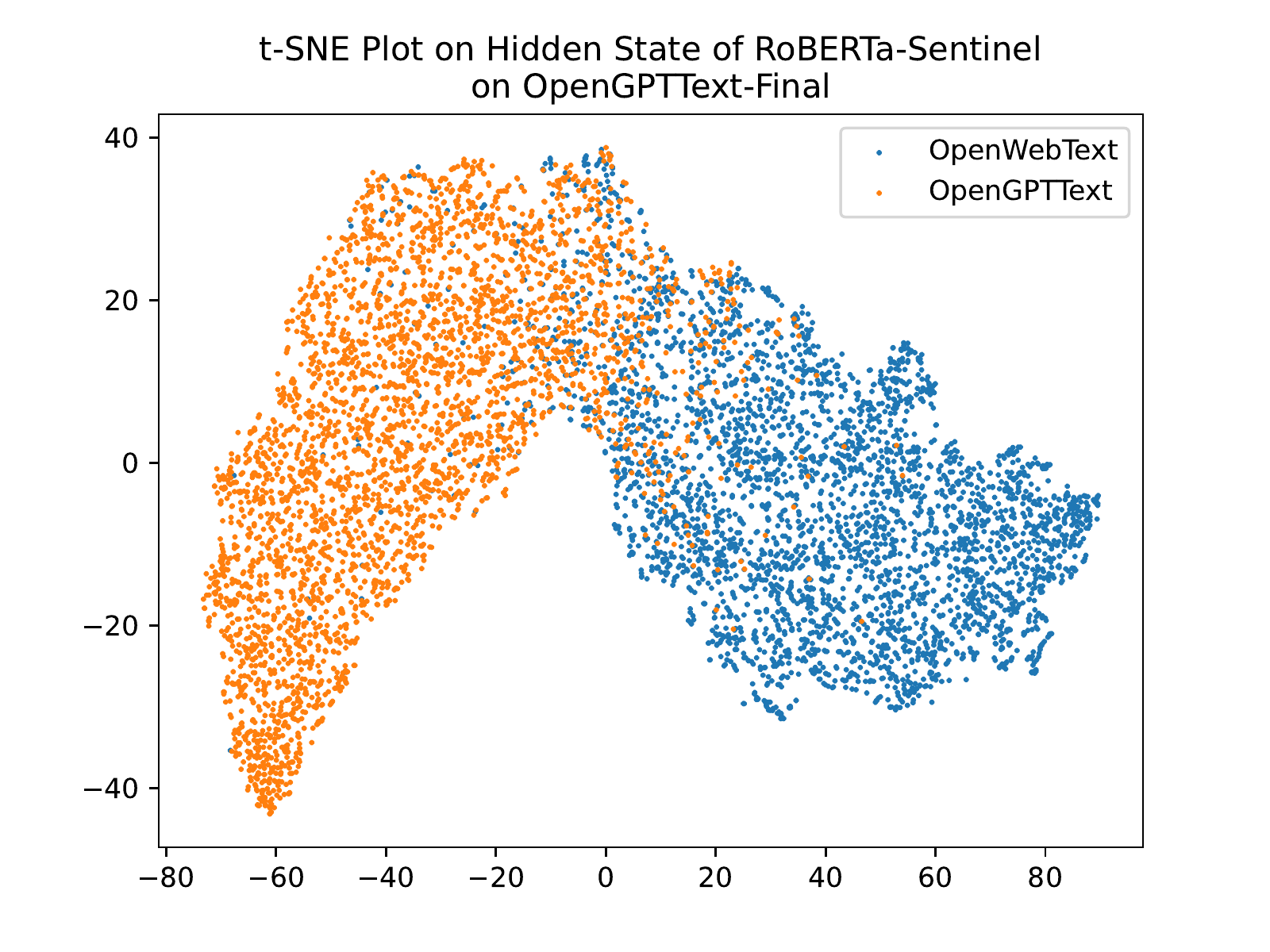}
    \caption{Hidden states of T5-Sentinel (Left) and RoBERTa-Sentinel (Right) on \texttt{urlsf-04} subset of \texttt{OpenGPTText-Final} after t-SNE dimensionality reduction.}
    \label{fig:t-sne result plot}
\end{figure}

\subsection{t-distributed Stochastic Neighbor Embedding Visualization}

t-distributed Stochastic Neighbor Embedding (t-SNE) projection proposed by Maatan et al. \cite{vanDerMaaten2008} is applied on the hidden state vector of both T5-Sentinel and RoBERTa-Sentinel. Results in figure \ref{fig:t-sne result plot} indicate that T5-Sentinel model can better separate the data set, which aligns with the performance of both models on test data set.

\section{Future Work}
Although our current model has shown promising results, there are certain limitations   in our current model. First and foremost, both T5-Sentinel and RoBERTa-Sentinel are trained with English corpus only. As a result, their performance on other languages such as Spanish or Chinese may not be optimal. To address this limitation, fine-tuning the models with non-English text can be helpful. However, it's worth noting that the pretrained version of the T5 model only supports English, French, Romanian, and German. Therefore, classification tasks involving languages other than these may require more than just fine-tuning alone.

In addition, the \texttt{OpenGPTText-Final} data set is collected with the prompt ``\texttt{Rephrase the following paragraph by paragraph}'', so the model trained on such data set might not perform well to other language tasks that ChatGPT is popularly used on, such as question answering or text generation. In the future, we plan to collect data sets involving a different textual context, like \texttt{eli5} \cite{eli5_lfqa} and \texttt{SQuAD} \cite{rajpurkar-etal-2016-squad}, to further assess the accuracy of the RoBERTa-Sentinel and T5-Sentinel on different tasks.

\section{Conclusion}
In conclusion, we have introduced a high-quality data set called \texttt{OpenGPTText}, which we have rephrased using the ChatGPT model. Additionally, we have designed, implemented, and trained two text classification models using RoBERTa and T5 architectures. Our models have achieved remarkable results, with accuracy exceeding 97\% on the test data set, as evaluated using various metrics.

Moreover, we have conducted an interpretability study to demonstrate our models' ability to extract and differentiate key features between human-written and ChatGPT-generated text. The study's results show that our models are effective in identifying the differences between the two types of text, providing insight into the strengths and limitations of the models and demonstrating their potential for real-world applications.

\section{Acknowledgement}
We would like to express our sincere appreciation to Professor Bhiksha Raj and our TA mentor Liangze (Josh) Li for their invaluable guidance, insightful comments, and constant support throughout the course of this research. Their expertise in the field have been instrumental in shaping our work. 

\bibliographystyle{unsrt} 
\bibliography{final_report}

\newpage
\appendix
\section{GPT2-Detector Baseline Analysis}

\subsection{Evaluation on GPT2-Baseline}

We reproduced the GPT2-Detector proposed by Solaiman et al. \cite{Solaiman2019} as the baseline model and performed evaluation on GPT2 output data set released by OpenAI \cite{GPT2-Output}. The results are shown below.

The GPT2-output data set contains random output from four variants of GPT2 model: small (117M parameter), medium (354M parameter), large (762M parameter) and extra-large (1542M parameter) and we have evaluate our baseline model on all of them. The result is shown in table \ref{tab:baseline reproduction evaluation}.

\begin{table}
  \caption{Baseline evaluation on GPT2-output data set}
  \label{tab:baseline reproduction evaluation}
  \centering
  \begin{tabular}{lcccccccc}
    \toprule
    & \multicolumn{2}{c}{Small} & \multicolumn{2}{c}{Medium} & \multicolumn{2}{c}{Large} & \multicolumn{2}{c}{Extra Large}\\
    \cmidrule{2-3} \cmidrule{4-5} \cmidrule{6-7} \cmidrule{8-9}
    Metrics        & top-$k$ & pure & top-$k$ & pure & top-$k$ & pure & top-$k$ & pure\\
    \midrule
    Accuracy       & $0.9648$ & $0.9623$ & $0.9627$ & $0.9567$ & $0.9616$ & $0.9449$ & $0.9498$ & $0.9310$\\
    False Positive & $0.0070$ & $0.0122$ & $0.0114$ & $0.0238$ & $0.0137$ & $0.0472$ & $0.0376$ & $0.0734$\\
    False Negative & $0.0319$ & $0.0319$ & $0.0319$ & $0.0319$ & $0.0319$ & $0.0319$ & $0.0319$ & $0.0319$\\
    \bottomrule
  \end{tabular}
\end{table}

The confusion matrices for every variant (small, medium, large, extra-large) and every sampling method (top-$k$ and pure) is shown in figure \ref{fig:Confusion Mats for baseline on GPT2 output data set, Pure} and figure \ref{fig:Confusion Mats for baseline on GPT2 output data set, top-k}.

\begin{figure}
    \centering
    \includegraphics[width=.8\textwidth]{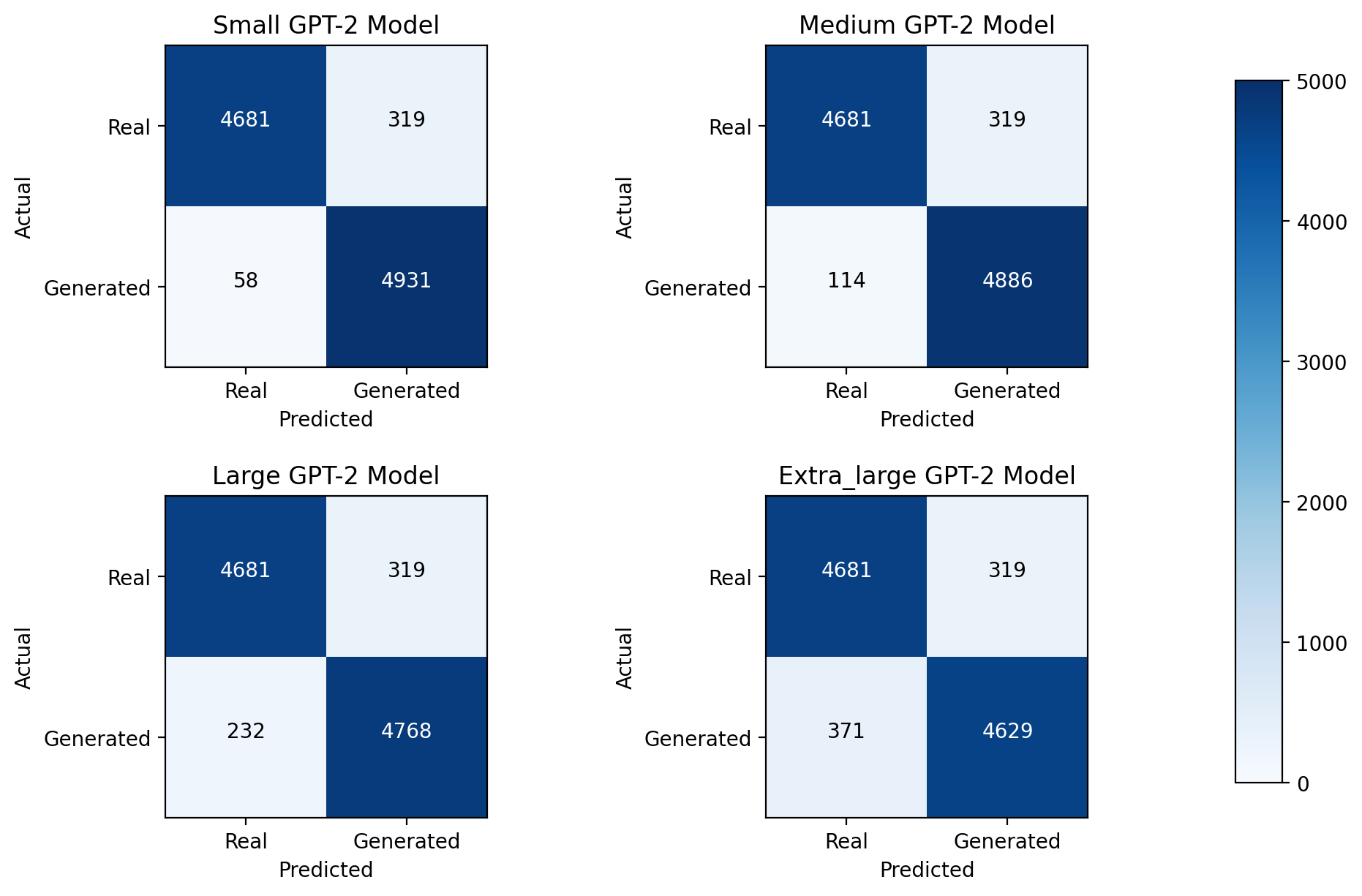}
    \caption{Confusion matrices of baseline model on GPT2-output data set under pure sampling method}
    \label{fig:Confusion Mats for baseline on GPT2 output data set, Pure}
\end{figure}

\begin{figure}
    \centering
    \includegraphics[width=.8\textwidth]{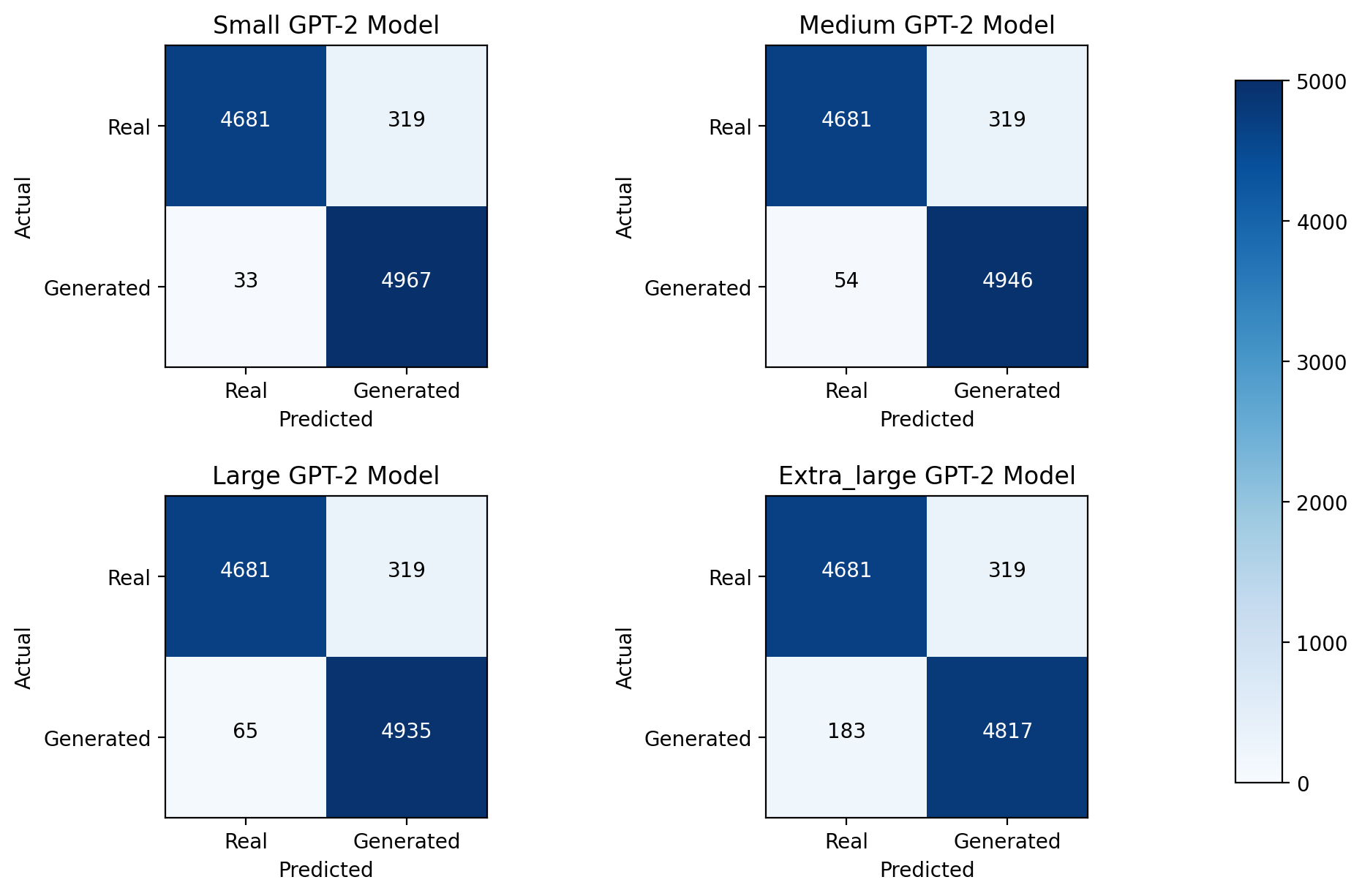}
    \caption{Confusion matrices of baseline model on GPT2-output data set under top-$k$ sampling ($k = 40$)}
    \label{fig:Confusion Mats for baseline on GPT2 output data set, top-k}
\end{figure}

\subsection{Trend Between Language Model Scale and Classification Accuracy}

\begin{figure}
    \centering
    \includegraphics[width=.98\textwidth]{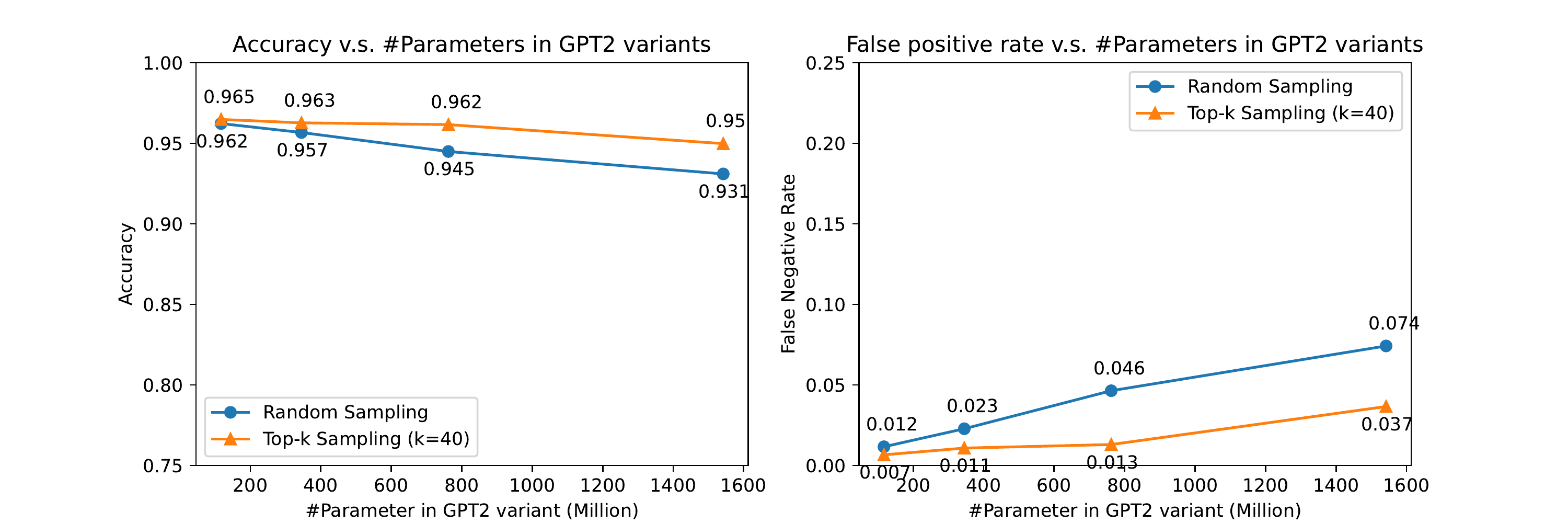}
    \caption{As the number of parameter in language model increase, the detector's accuracy decreases linearly, while the false positive  rate increases.}
    \label{fig:Trend of acc in GPT2-Output dataset (Appendix)}
\end{figure}

When running baseline test of GPT2-Detector, we noticed that there is an approximately linear relationship between classification accuracy and the scale of language model, as illustratd in \ref{fig:Trend of acc in GPT2-Output dataset (Appendix)}.

\subsection{PCA Analysis for Baseline Model}

Despite the baseline model's strong performance in detecting GPT-2 generated content, it encounters significant challenges when tasked with detecting GPT-3.5 generated content. In fact, without any additional training, the baseline model achieved a mere 54.98\% accuracy on the OpenGPTText dataset, only slightly better than random chance. This significant drop in accuracy highlights the challenges of differentiating between human-generated and GPT-3.5 generated content, likely due to the increased complexity of the GPT-3.5 model.

\begin{figure}
    \centering
    \begin{tabular}{cc}
          \includegraphics[width=.45\textwidth]{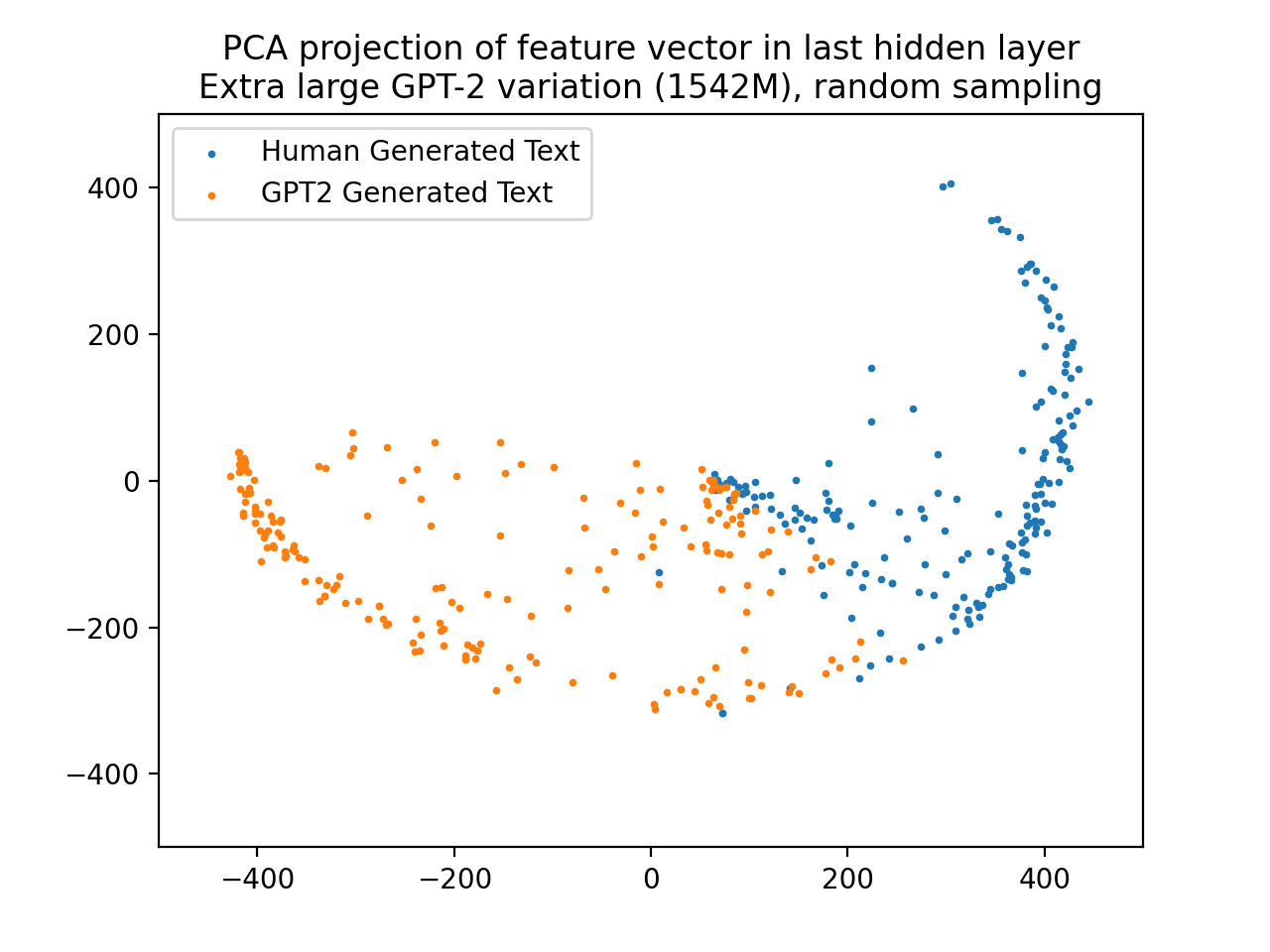}
          & 
          \includegraphics[width=.45\textwidth]{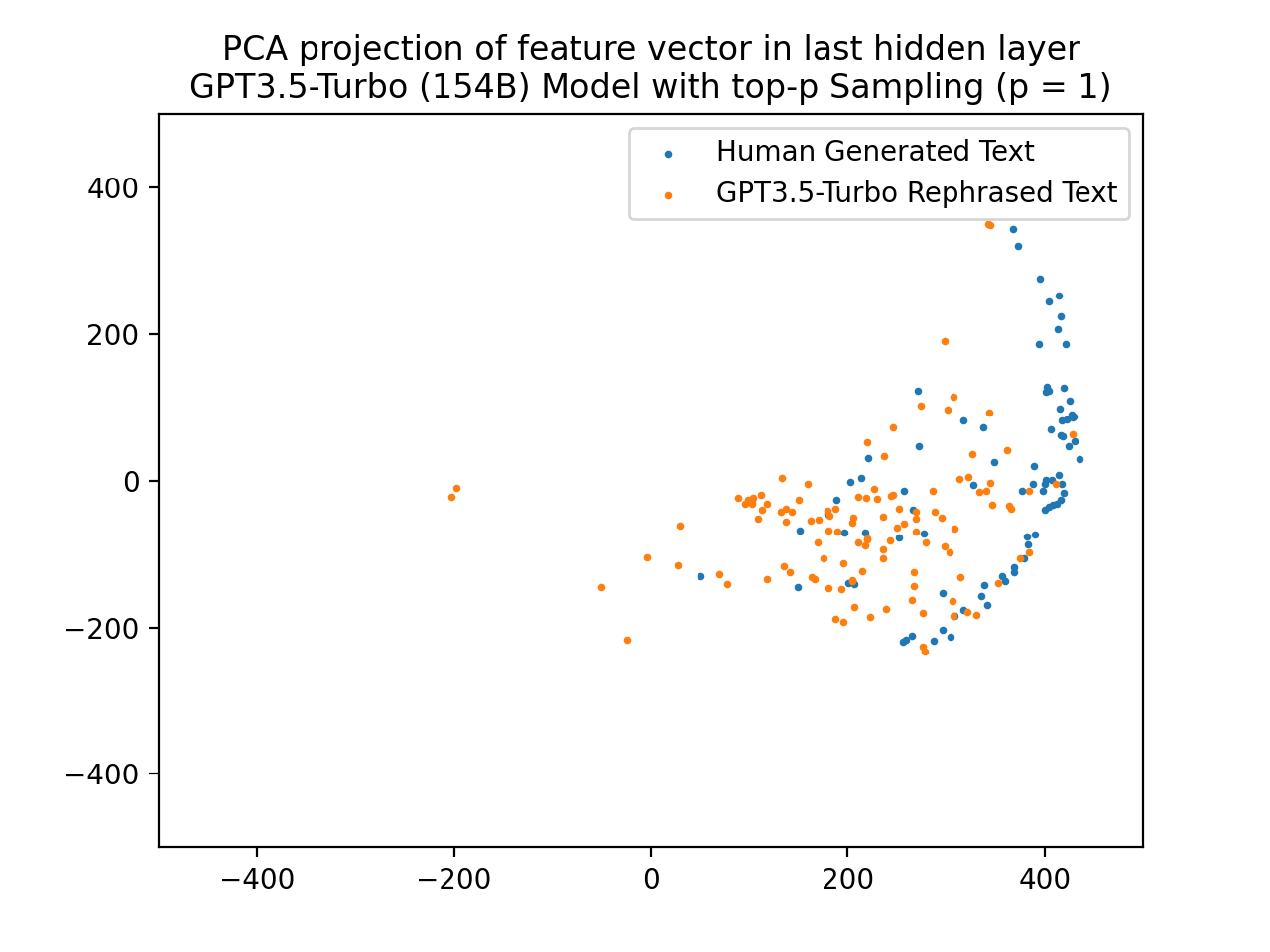}
    \end{tabular}
    \caption{PCA Projection: GPT-2 vs. GPT-3.5 Turbo}
    \label{fig:PCA Projection: GPT-2 vs. GPT-3.5}
\end{figure}

Another notable observation is the difference in PCA projections between GPT-2 and GPT-3.5 generated content. The PCA projection for GPT-2 indicates that human-generated and GPT-2 generated content are clearly distinguishable from each other. However, the same distinction is not as clear in the GPT-3.5 projection, as shown in figure \ref{fig:PCA Projection: GPT-2 vs. GPT-3.5}.

\newpage
\section{Detailed Information for Evaluation}

\subsection{Evaluation Result}

As shown in table \ref{tab:baseline reproduction evaluation}. The metrics are calculated under the positive probability threshold of $0.5$.

\begin{table}
    \caption{Evaluation Result on OpenGPTText-Final (Row 1-3), OpenGPTText-Original (Row 4-7), and GPT2-Output (Row 8-11). TPR stands for ``True Positive Rate'', TPC stands for ``True Positive Count'', TNR stands for ``True Negative Rate'', TNC stands for ``True Negative Count''.}
    \centering
    \begin{tabular}{lccccc}
    \toprule
    Model & Accuracy & TPR, (TPC) & TNR, (TNC) & FPR, (FPC) & FNR, (FNC) \\
    \midrule
    T5      & 97.98\% &98.71\%, (2906) & 97.25\%, (2863) & 2.75\%, (81) & 1.29\%, (38) \\
    RoBERTa & 93.92\% &96.81\%, (2850) & 91.03\%, (2680) & 8.97\%, (264) & 3.19\%, (94) \\
    OpenAI  & 57.68\% &20.24\%, (596) & 95.11\%, (2800) & 4.89\%, (144) & 79.76\%, (2348) \\
    ZeroGPT & 54.36\% &34.99\%, (1030) & 73.74\%, (2171) & 26.26\%, (773) & 65.01\%, (1914) \\
    GPT2    & 38.5\% &13.21\%, (389) & 97.16\%, (1233) & 2.84\%, (36) & 86.79\%, (2555) \\
    \midrule
    T5      & 97.64\% &98.74\%, (2907) & 96.54\%, (2842) & 3.46\%, (102) & 1.26\%, (37) \\
    RoBERTa & 88.28\% &98.13\%, (2889) & 78.43\%, (2309) & 21.57\%, (635) & 1.87\%, (55) \\
    OpenAI  & 56.64\% &14.84\%, (437) & 98.44\%, (2898) & 1.56\%, (46) & 85.16\%, (2507) \\
    ZeroGPT & 56.1\% &28.67\%, (844) & 83.53\%, (2459) & 16.47\%, (485) & 71.33\%, (2100) \\
    GPT2    & 37.86\% &12.84\%, (378) & 95.9\%, (1217) & 4.1\%, (52) & 87.16\%, (2566) \\
    \midrule
    T5      & 48.68\% &3.3\%, (165) & 94.06\%, (4703) & 5.94\%, (297) & 96.7\%, (4835) \\
    RoBERTa & 46.56\% &10.36\%, (518) & 82.76\%, (4138) & 17.24\%, (862) & 89.64\%, (4482) \\
    OpenAI  & 71.22\% &56.02\%, (2801) & 86.42\%, (4321) & 13.58\%, (679) & 43.98\%, (2199) \\
    ZeroGPT\tablefootnote{ZeroGPT failed to process two data entries (with ID: 255332 and 258673 in \texttt{xl-1542M.test.jsonl}) in the GPT2-Output data set, those two entries are not counted in the calculation of metrics.} 
    & 43.09\% &9.52\%, (476) & 76.64\%, (3832) & 23.36\%, (1168) & 90.48\%, (4522) \\
    GPT2    & 93.1\% &92.58\%, (4629) & 93.62\%, (4681) & 6.38\%, (319) & 7.42\%, (371) \\
    \bottomrule
    \end{tabular}
    \label{tab:Detailed Evaluation Data (Appendix)}
\end{table}

\subsection{Content Sample}

\subsubsection{Sample 1 - [urlsf\_subset04]-[236996]-web}

\begin{lstlisting}[breaklines=true, breakindent=1em, breakautoindent=true, basicstyle=\ttfamily\small]
Lexus IS 250 is the entry model among IS sedans. Pictured is the optional F Sport package that firms already stiff suspension and adds some distinguishing visuals. (Photo11: Lexus)
Lexus gave us enough go-fast imagery in its Super Bowl ads last Sunday that it almost seemed to be saying, "See, see, are too sporty."
And, yes, some malingerers still might be insisting, "Are not," and need nudging.
After all, Lexus' reputation for years has been luxury at the expense of handling and performance. But the brand is trying to transform so that you'll compare it with BMW instead of Buick.
The ES, at the lower end, and the LS, at the top, still could be considered luxo-machines more than yippee-mobiles. But most Lexus models we've driven lately behave in sporty fashion when prodded that direction by the driver.
(...Truncated)
\end{lstlisting}

\subsubsection{Sample 2 - [urlsf\_subset04]-[246672]-gpt}
\begin{lstlisting}[breaklines=true, breakindent=1em, breakautoindent=true, basicstyle=\ttfamily\small]
After succumbing to a hip injury, Nick Kyrgios has come to the decision to take rest and heal. This has led him to reflect on 2017, with its highs and lows, from consecutive wins against great tennis players to the difficulties he faced at Grand Slams. However, his fondest memories were from team events, particularly the Laver Cup and the Davis Cup. Kyrgios reveals that tennis can be a lonely sport, and he often struggles with it. However, he praises the team spirit in Davis Cup, highlighting how they all support each other, win or lose, and says it made him feel like he was part of the team. 
Rusty, Davis Cup captain, created a WhatsApp group a year and a half ago, which includes all the Davis Cup players. Kyrgios recounts how, after his semi-final win in Beijing, his phone was flooded with loyal messages from the coaches and players. This sparked his realization that it had become bigger than tennis, and they had become a family trying to keep in touch no matter how far apart they all were. He feels it has helped provide a support system for everyone, both on and off the court, at a time when they really needed it the most. 
(... Truncated)
\end{lstlisting}

\subsubsection{Sample 3 - [urlsf\_subset04]-[230559]-web}
\begin{lstlisting}[breaklines=true, breakindent=1em, breakautoindent=true, basicstyle=\ttfamily\small]
Add DevOps Tools to your Pipeline
Densify XL Impact Bitbucket Bitbucket Server Bower Crucible Deveo Fisheye Gerrit Git GitHub GitLab Gogs Helix ISPW Kallithea Mercurial Micro Focus AccuRev Micro Focus StarTeam Perforce HelixCore Rational Clearcase Rational Team Concert Subversion Team Foundation Server Team Foundation Version Control

(... Truncated)

Kubernetes Engine Kubernetes Linux Containers Marathon Mesos Mesosphere DC/OS Nomad OpenVZ Portainer Rancher Solaris Containers Supergiant Swarm Sysdig Tectonic Weaveworks rkt OpsGenie DBmaestro Datical Delphix Flyway Idera Liquibase Quest Toad Redgate Redgate SQL Toolbelt
Add tools from the Periodic Table of DevOps or select from the full list above.
Click "Visualize My Pipeline!" to view your pipeline in the DevOps Diagram Generator.
\end{lstlisting}

\subsubsection{Sample 4 - [urlsf\_subset04]-[313139]-web}
\begin{lstlisting}[breaklines=true, breakindent=1em, breakautoindent=true, basicstyle=\ttfamily\small]
A carnival reveller dressed as a clown celebrates on the street in Berlin February 18, 2007. REUTERS/Pawel Kopczynski
LONDON (Reuters) - Bad news for Coco and Blinko -- children don't like clowns and even older kids are scared of them.
The news that will no doubt have clowns shedding tears was revealed in a poll of youngsters by researchers from the University of Sheffield who were examining how to improve the decor of hospital children's wards.
The study, reported in the Nursing Standard magazine, found all the 250 patients aged between four and 16 they quizzed disliked the use of clowns, with even the older ones finding them scary.
"As adults we make assumptions about what works for children," said Penny Curtis, a senior lecturer in research at the university.
"We found that clowns are universally disliked by children. Some found them quite frightening and unknowable."
(End of File)
\end{lstlisting}

\end{document}